\documentclass{article}

\usepackage{arxiv}

\usepackage[utf8]{inputenc}
\usepackage[T1]{fontenc}
\usepackage{hyperref}
\usepackage{url}
\usepackage{booktabs,tabularx,array,multirow,longtable}
\usepackage{amsmath,amssymb,amsfonts,amsthm}
\usepackage{nicefrac}
\usepackage{microtype}
\usepackage{graphicx}
\usepackage[dvipsnames]{xcolor}
\usepackage[numbers,sort&compress]{natbib}
\usepackage{doi}
\usepackage{authblk}

\graphicspath{{figures/}}
\newcolumntype{Y}{>{\raggedright\arraybackslash}X}
\newcommand{\figref}[1]{Fig.~\ref{#1}}
\newcommand{\hl}[1]{{\setlength{\fboxsep}{0.8pt}\colorbox{orange!40}{#1}}}
\newcommand{\hm}[1]{{\setlength{\fboxsep}{0.8pt}\colorbox{teal!25}{#1}}}
\newcommand{\RQ}[1]{\textbf{RQ#1}}
\newcommand{\Description}[2][]{}
\theoremstyle{definition}
\newtheorem{definition}{Definition}[section]
\hypersetup{colorlinks=true,linkcolor=MidnightBlue,citecolor=MidnightBlue,urlcolor=MidnightBlue}

\title{From Opinions to Commitments: Unstable Assignments and Silent Choices in AI-Mediated Proposals}

\author{Sachit Mahajan\thanks{Corresponding author: \texttt{smahajan@ethz.ch}}}
\author{Javier Argota S{\'a}nchez-Vaquerizo}
\author{Arthur Capozzi}
\author{Dirk Helbing}
\affil{Computational Social Science, ETH Zurich, Zurich, Switzerland}
\date{}

\renewcommand{\shorttitle}{From Opinions to Commitments}

\hypersetup{
pdftitle={From Opinions to Commitments: Unstable Assignments and Silent Choices in AI-Mediated Proposals},
pdfauthor={Sachit Mahajan, Javier Argota S\'anchez-Vaquerizo, Arthur Capozzi, Dirk Helbing},
pdfkeywords={AI mediation, digital democracy, human agency, pluralism, design for values, transparency, reproducibility, algorithm audit},
}

\begin{document}
\maketitle

\begin{abstract}
Large language models (LLMs) are increasingly being used to turn citizens' opinions into collective proposals. These proposals can make assignments: who should act, what must be done, how much one should spend and by when. However, if the assignments change when the same opinions are processed again, publishing a single proposal can hide alternatives that participants never get to see. We examined this issue in four LLM mediators (three open-weight and one commercial) using opinions from DeliberationBank and Polis. For each task, a mediator drafted a proposal from thirty opinions five times, varying either the random seed or the order of the opinions. In 17\% of five-run sets from the open-weight mediators and 20\% from the commercial one, some runs contained assignments while others did not. Reordering the opinions changed the assignments almost as often as replacing them with opinions from different citizens. Three human annotators confirmed that the detected differences between runs were real, although the LLM judge overstated how often they occurred. Substantial variation remained under four safeguards: temperature zero, a commercial mediator, draft selection, and draft merging. The mediator's objective also shaped representation: prompts seeking agreement covered minority positions much less often than prompts asking to describe differences. A single proposal can therefore obscure choices in both the assignments it proposes and the views it excludes. We propose audits that make alternative assignments visible and disclose the objective behind each proposal.

\end{abstract}


\section{Introduction}
\label{sec:intro}

Participatory budgeting, citizen assemblies, and online consultations give citizens a direct say in public decisions \citep{cabannes2004participatory,fishkin2005experimenting,smith2009democratic}. Platforms such as Polis, vTaiwan and Decidim now involve thousands of people at once \citep{small2021polis,hsiao2018vtaiwan,barandiaran2024decidim}. The difficult step has therefore moved from collecting input to making sense of it. Large language models (LLMs) are increasingly used to perform this step. They write consensus statements that groups prefer over those by human mediators \citep{bakker2022finetuning,tessler2024ai}, summarize Polis conversations \citep{small2023opportunities}, draft policy guidelines from collective dialogues \citep{konya2023democratic}, and analyze responses to national consultations \citep{iai2025consult}. Other systems, such as CIME and the Habermas Machine, use AI to support structured group deliberation and synthesize diverse perspectives through iterative feedback and voting \citep{tessler2024ai, capozzi2026cime}. Existing evaluations have examined factors such as efficiency, alignment with human analysts, and acceptance of AI-generated outputs by participants \citep{iai2025consult,bakker2022finetuning}. However, two additional aspects deserve attention: variability in proposals generated from the same opinions, and the influence of mediation objectives on which views are represented. As these systems move from summarizing views to drafting proposals, the assignment of responsibilities may become part of the decision-making process \citep{mahajan2026participatory}, which makes the design of mediation essential for democratic participation \citep{landemore2024can,friedman2019value,vandenhoven2015design}.

\begin{figure}[t]
  \centering
  \includegraphics[width=\textwidth]{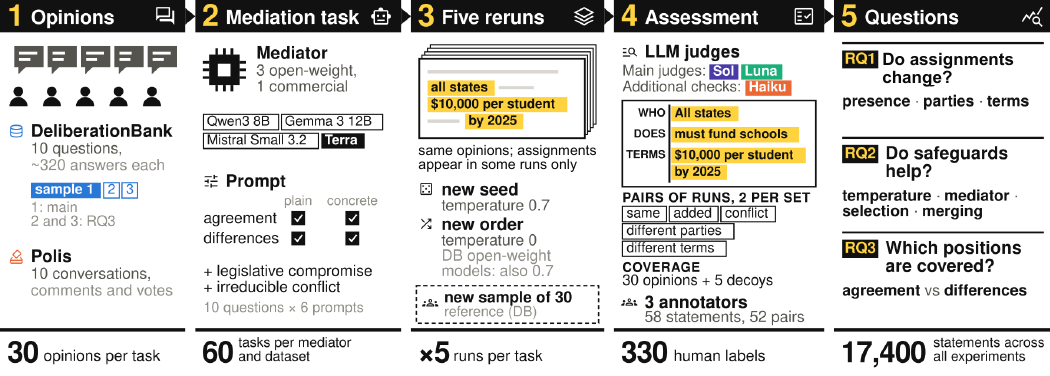}
  \Description{A left-to-right schematic: citizens with speech bubbles, an AI mediator chip with two objectives, a fanned stack of five statement cards with differing highlighted assignments and a dashed reference card for other citizens' opinions, a magnifier for the LLM judge with three annotator icons, three tiles for the research questions, and a footer with the total number of statements.}
  \caption{\textbf{Study design.} Each task combines thirty opinions, a mediator and one of six prompts. Five runs vary sampling or opinion order; Terra uses independent sampling calls rather than specified seeds. New opinion samples provide a DeliberationBank reference. Judges assess assignments and coverage. Three annotators check assignments on 58 statements and 52 pairs, including six controls (330 judgments). The 17{,}400 statements cover all experiments (Appendix~\ref{app:ledger}).}
  \label{fig:overview}
\end{figure}

\figref{fig:overview} outlines our audit of assignment stability, procedural safeguards and the representation of citizens' positions.

Existing evaluations examine whether participants endorse a statement and whether it represents their views \citep{tessler2024ai,zhu2025can,zhang2024fair}. A further question concerns the alternatives that a single output leaves unseen. In \figref{fig:example}, the same model received the same thirty opinions in two different orders at temperature zero. The first output recommends policies without explicit assignment of responsibilities to a particular party. The second calls on every state to spend at least \$10,000 per student by 2025. If an organizer publishes only one run, readers cannot see that changing the input order produces a different assignment of responsibilities. We call this a \emph{silent choice}: a procedural factor that carries no information about citizens' views changes the assignments in a proposal, while the published output leaves the alternative undisclosed.

\begin{figure}[!htb]
\centering\small
\setlength{\fboxsep}{6pt}
\fbox{\begin{minipage}[t]{0.465\textwidth}
\textbf{Run A} \hfill \emph{opinions in order A, $T=0$}\par\smallskip
Citizens agree that education is vital for the future of the country, and that funding should be allocated responsibly to support teaching, research, and student access. They can both accept that academic institutions should receive stable and adequate funding, with clear guidelines on how funds are distributed, ensuring that all students, regardless of background, have access to quality education. Funding should be determined through transparent processes involving educators, experts, and community representatives, with strict oversight to prevent misuse and ensure that cuts do not compromise essential research or student support.\par\smallskip
\textit{Assigned parties:} none. \textit{Specified terms:} none.
\end{minipage}}\hfill
\fbox{\begin{minipage}[t]{0.465\textwidth}
\textbf{Run B} \hfill \emph{same opinions in order B, $T=0$}\par\smallskip
Citizens agree that education is a critical foundation for societal and economic success, and that funding should be prioritized to support students, researchers, and institutions. To ensure fairness and effectiveness, \hl{all states} should allocate \hl{at least \$10,000 per student annually} from existing federal budgets \hl{by 2025}, with transparent oversight to prevent misuse, and ensure that low-income and minority students receive equitable access to resources.\par\smallskip
\textit{Assigned parties:} all states. \textit{Specified terms:} at least \$10,000 per student annually; by 2025.
\end{minipage}}
\caption{\textbf{Two runs on the same opinions, but presented in a different order.} Qwen3 8B received the same thirty DeliberationBank opinions on cuts to academic funding, with the concrete agreement prompt at temperature zero; only their order differed. Run~A names no party that must act, while Run~B requires all states to meet an amount and a deadline (highlighted), although none of the thirty opinions mentions an amount or a deadline. The statements are verbatim; we chose this pair as an illustration.}
\Description{Two boxes side by side with the text of two statements from the same task; the second contains a specific assignment for all states with an amount and a deadline, highlighted, while the first recommends policies without naming an assigned party.}
\label{fig:example}
\end{figure}

Such choices matter most when mediation is used to draft policy. A proposal that names who must act, what must be done, how much and by when is the part of mediation that decision-makers can adopt, and it is where pluralism and human agency are most directly affected \citep{helbing2019will,weyl2024plurality}. It is therefore important to know how far these proposed assignments follow from what citizens said. Other choices are made deliberately and affect the outcome as well. Organizers decide whether the mediator should look for agreement among citizens or set out where they differ. This choice is in the prompt, but its effects are not visible: a prompt may change whose positions remain in the proposal, while asking to highlight agreements may present a consensus that citizens have not reached. Therefore, three research questions guide our study:
\begin{itemize}
  \item[\RQ1] When an AI mediator is run again on the same opinions, does it propose the same assignments? How large is this variation compared to the variation that arises when different citizens are consulted?
  \item[\RQ2] Do common procedural safeguards (temperature zero, another mediator, selecting one draft or merging several drafts) make the assignments stable?
  \item[\RQ3] Does the objective given to the LLM mediator change whose positions the proposal includes?
\end{itemize}

To answer these questions, we ran four LLM mediators on real opinions from DeliberationBank and Polis, repeated every task under controlled changes, and checked the assignment measures against three human annotators. We used LLM judges only as a research tool to measure what the statements said. We do not recommend them for evaluating citizens' opinions or choosing between proposals in a participatory process, as these decisions belong to the participants, and handing them over to a model would take away their agency~\citep{helbing2019will}. Our paper makes three contributions. First, it defines silent choices and relates them to human agency, pluralism, transparency, equal treatment and contestability (Section~\ref{sec:concept}). Second, it describes a rerun audit that needs only query access to a mediator, while assignment measures are checked against human readers (Section~\ref{sec:design}). Third, it reports evidence on the three research questions based on two datasets (Section~\ref{sec:results}) and derives requirements for AI-supported participation that organizers and researchers can test (Section~\ref{sec:discussion}).

\section{Related work}
\label{sec:related}

\emph{Evaluating AI mediation.} Research on AI mediation has established that LLMs can write statements people accept. \citet{bakker2022finetuning} fine-tuned a model to generate consensus statements and selected candidates by predicted group approval. The Habermas Machine added rounds of human critique, and participants preferred LLM statements to those of human mediators \citep{tessler2024ai}. Later work asked whose views a statement keeps. DeliberationBank documents minority under-representation across eighteen LLM summarizers and finds no consistent effect of an opinion's position in the input on its representation \citep{zhu2025can}. Generative social choice seeks representation guarantees for selected statements \citep{fish2026generative}, and Habermolt reports a trade-off between representativeness and actionability \citep{low2026habermolt}. Related studies examine disagreement between models in the analysis of public comments \citep{najera2026when}, repeated LLM votes in participatory budgeting \citep{majumdar2026generative} and prompt injection \citep{gudinorosero2025prompt}. Appendix Table~\ref{tab:literature} summarizes this work. This literature examines endorsement, representation and several sources of variation. Our focus is the stability of explicit assignments when a mediator processes unchanged opinions, together with the effect of the mediation objective on position coverage.

\emph{Arbitrariness, accountability and design for values.} Models that are equally accurate can decide differently about the same person \citep{marx2020predictive,black2022model,mahajan2026revisiting}. Such arbitrariness raises moral concerns \citep{creel2022algorithmic} and can be measured through consistency across replications \citep{cooper2024arbitrariness}. LLM outputs vary with prompt format \citep{sclar2024quantifying}, with the position of information in the input \citep{liu2024lost} and with inference conditions, even at temperature zero \citep{atil2025nondeterminism,ouyang2025nondeterminism,he2025defeating}. Randomized procedures can be legitimate when their operation is disclosed and bounded \citep{chooi2026finding}. Design for values asks how systems can meet requirements of transparency, accountability and autonomy \citep{friedman2019value,vandenhoven2015design}; in digital democracy these include preserving diversity and supporting self-organization \citep{helbing2019will,helbing2023democracy,page2007difference}. Accountability requires opportunities to trace, explain and contest decisions \citep{kroll2017accountable,wieringa2020account,alfrink2023contestable}, supported by audits and provenance records \citep{raji2020closing,mokander2024auditing,moreau2013prov,zaghouani2026accountable,mahajan2026participatory}. When an LLM settles a contested meaning through one interpretation \citep{gurarieh2026ambiguity}, access to the alternatives becomes part of what contestation requires \citep{stewart2026evidentiary}. Without such access, participation can turn into a legitimization ritual \citep{sloane2022participation,birhane2022power}.

\emph{LLM judges as instruments.} LLM judges make large-scale measurement possible \citep{zheng2023judging,gilardi2023chatgpt,ziems2024large}. Their agreement with human coders varies with tasks \citep{tornberg2024large,bavaresco2025llms,thakur2025judging}, and they may favor outputs from their own model family \citep{panickssery2024llm}. We therefore checked the assignment measures against human readings, relied on human labels to establish differences between assignments \citep{egami2023using,angelopoulos2023prediction} and repeated key analyses with a judge from another developer.

\section{Silent choices}
\label{sec:concept}

\subsection{Why stable assignments matter}

If reordering the same opinions changes a proposal's assignments, presentation order shapes the proposal without carrying information about citizens' views. This affects five values (Table~\ref{tab:values}). \emph{Human agency} requires that participants can consider alternatives before choosing which assignments to support and turn into commitments \citep{helbing2019will}. \emph{Pluralism} keeps diverse and minority positions visible, because the range of perspectives and solution proposals gives collective intelligence its capacity to find better solutions \citep{page2007difference,weyl2024plurality}. \emph{Transparency} requires records of how drafts were generated and selected \citep{chooi2026finding}; \emph{equal treatment} requires scrutiny when input order changes who is assigned \citep{creel2022algorithmic}; and \emph{contestability} requires opportunities to question assignments and seek revision \citep{pettit1997republicanism,alfrink2023contestable}. %

These concerns do not require identical wording or rule out multiple drafts. New evidence can justify different assignments, and alternatives can enrich deliberation when participants can examine them. The problem arises when variation stays hidden behind one published proposal \citep{chooi2026finding}. The same holds for the mediation objective: a generated statement of agreement does not show that participants agree, so the objective should be stated and justified.  %

\subsection{Definitions}

\begin{definition}[Mediation task and reruns]
A \emph{mediation task} $\tau=(q,C,m,p)$ consists of a public question $q$, an unordered collection $C$ of thirty opinions on it, a mediator model $m$ and a prompt $p$. Running the task produces a statement. A \emph{rerun set} contains five statements $s_1,\dots,s_5$ produced from the same task, while one factor changes: the random seed (\emph{seed reruns}) or the order in which the opinions in $C$ are presented (\emph{order reruns}). Neither factor carries information about what citizens think. For comparison, \emph{new-sample reruns} replace $C$ with five separate collections of thirty opinions on the same question, so that the evidence itself changes.
\end{definition}

\begin{definition}[Assignments]
We define an \emph{assignment} as a proposed requirement directed toward an identifiable party. Each \emph{assignment} specifies who should act (the \emph{bearer} $b$), what they should do (the \emph{action} $a$), and, when applicable, additional terms ($\theta$), such as amounts, deadlines, thresholds, eligibility rules or sanctions \citep{wyner2011rule,kiyavitskaya2008automating}. In \figref{fig:example}, run~B contains the assignment $(b,a,\theta)=$ (all states, allocate funding, at least \$10,000 per student, by 2025). Vague collectives (``we'', ``society'') and abstract nouns are not bearers, and a requirement counts only if the statement itself proposes it: a statement that reports what some citizens believe should happen assigns no one. We call the set of assignments in a statement its \emph{assignment frame}, write $B(s)$ for the set of bearers, and say that $s$ \emph{assigns someone} if $B(s)\neq\emptyset$. We use \emph{commitments} for assignments that participants explicitly endorse and the assigned parties accept. Our audit measures proposed assignments in the text and makes no claim about their acceptance or legal enforceability. Whether a proposal should contain such assignments or not is a separate question.
\end{definition}

\begin{definition}[Mixed sets and party agreement]
A rerun set is \emph{mixed} if some but not all of its five statements assign someone. For rerun sets in which all five statements assign someone, the \emph{party agreement} $R(\tau)$ is the mean Jaccard similarity of the bearer sets over the ten pairs of runs:
\begin{equation}\label{eq:R}
R(\tau)=\frac{1}{10}\sum_{1\le i<j\le 5} J\big(B(s_i),B(s_j)\big),\qquad J(X,Y)=\frac{|X\cap Y|}{|X\cup Y|}.
\end{equation}
\end{definition}

The Jaccard similarity is the share of all parties named in either of two runs that are named in both (1, the same parties; 0, none shared). It handles runs that name different numbers of parties and is the set-based analogue of the self-consistency measures used to study arbitrariness across replications \citep{cooper2024arbitrariness}. We compute it only for rerun sets in which every run assigns someone, because two runs that assign no one would otherwise count as perfect agreement.

\begin{definition}[Differing assignments]
When at least one of two runs assigns someone, the pair falls into one of five relations \citep{maccartney2008modeling}: \emph{same assignments} (only the wording differs); \emph{added assignment} (one run contains every assignment of the other and at least one more, without conflict); \emph{different parties}; \emph{different terms} (the same parties with different amounts, deadlines, thresholds, eligibility rules or sanctions); and \emph{conflicting assignments} (the two sets of requirements cannot both be adopted). When a pair shows multiple relations, we classify it by the most serious one in the following order: conflicting, different parties, different terms, added, same. Two runs have \emph{differing assignments} if the relation is \emph{different parties}, \emph{different terms} or \emph{conflicting assignments}, because adopting one run instead of the other would change who has to act or on which terms.
For a set of compared pairs,
\begin{equation}\label{eq:M}
D=\frac{\text{pairs with differing assignments}}{\text{pairs in which at least one run assigns someone}}.
\end{equation}
$D$ counts pairs: a pair counts once, no matter how many of its assignments differ.
Compatible additions are not counted as \emph{differing assignments}, because they do not change or contradict assignments already present in the other run.
How much the assigned parties overlap is measured separately by party agreement $R$ (Eq.~\ref{eq:R}).
\end{definition}

A \emph{silent choice} occurs when a mediator produces different assignments from the same opinions, but participants are shown only one proposal as the collective conclusion, with the alternatives left undisclosed. Our audit measures the variation across reruns that could give rise to a silent choice. 

\section{Study design}
\label{sec:design}

\subsection{Opinions, mediators and prompts}
\label{sec:data}

\emph{Opinions.} DeliberationBank \citep{zhu2025can} contains free-text answers from US adults to ten policy questions, for example whether the government should provide basic health insurance for everyone (Appendix~\ref{app:data}). It has about 320 answers per question, so several separate samples of thirty opinions can be drawn for each question. The main analyses use one sample of thirty opinions per question (sample~1); RQ3 also uses two further non-overlapping samples (samples~2 and~3). Polis \citep{small2021polis} provides records of online consultations, in which participants write short comments and vote on the comments of others. We used the ten English-language policy conversations in the Polis openData repository with at least thirty eligible comments (Appendix~\ref{app:data}). Each task gives the mediator thirty opinions, a group comparable to a small citizen panel or a breakout group in a larger consultation. We selected thirty opinions per task to allow multiple non-overlapping samples per DeliberationBank question and to make it feasible for judges to assess whether each opinion was covered. All mediators were instructed to write statements of two to four sentences.

\emph{Mediators.} The four main mediators are three open-weight models, whose parameters are publicly available, from different developers (Qwen3 8B, Gemma 3 12B and Mistral Small 3.2 24B), and one commercial model, GPT-5.6 Terra. For DeliberationBank's samples~2 and~3, the statements were written by Qwen3 8B, Gemma 3 12B, Llama 3.1 8B and Phi-4 14B (all models and roles are listed in Appendix Table~\ref{tab:models}). We used two decoding settings: temperature zero, which always selects the most likely next token, and temperature 0.7, which samples tokens with some randomness controlled by a random seed.

\emph{Prompts.} Six prompts tell the mediator what to write, and we refer to them by short names throughout. Four prompts combine an objective with a style. The objective is either to \emph{find agreement} (``set out what these citizens can agree on'') or to \emph{set out differences} (``set out where these citizens differ''). The style is either plain or concrete. The concrete version replaces the plain-language instruction with ``say who pays, how much, who qualifies, what happens if the rule is broken, and by when''. These four prompts are called \emph{agreement}, \emph{concrete agreement}, \emph{differences}, and \emph{concrete differences}. Two further prompts follow earlier approaches to mediation. The \emph{legislative compromise} prompt asks the mediator to act as an arbitrator who drafts an operational compromise between factions, with concrete mechanisms, budgets, and enforcement. The \emph{irreducible conflict} prompt asks it to state the conflicts that cannot be resolved and to propose rules of engagement, without forcing consensus. Appendix~\ref{app:prompts} describes the prompt design; all six full prompts are included with the reproduction code.\footnote{Data and code for reproducing the reported results are available through an \href{https://osf.io/42krw/overview?view_only=d9e75914e7d7412c88b0bfe01bb6e3f6}{OSF repository}.} The concrete and legislative prompts reflect the use of mediation for drafting policy \citep{konya2023democratic,low2026habermolt}.

\subsection{Comparisons}

\emph{Stability (RQ1).} For every main mediator, dataset, prompt and question, we produced five reruns with five different random seeds (at temperature 0.7) and five reruns with different orders of the same opinions (at temperature zero; on DeliberationBank also at temperature 0.7). This gave 60 tasks (ten questions $\times$ six prompts) per mediator, dataset, and type of rerun. Terra's interface did not accept a seed, so its seed reruns were five independent samples at temperature 0.7. For the three open-weight mediators on DeliberationBank, we also generated proposals from five new samples of thirty opinions on each question. We compared these with order reruns of the original opinions to assess how much proposals change when the opinions themselves change. Polis had too few comments per conversation for five separate samples.

\emph{Safeguards (RQ2).} We tested four ways an organizer might try to reduce variation:
\begin{itemize}
    \item \emph{Temperature zero:} disable random token sampling.
    \item \emph{Commercial mediator:} use Terra in place of an open-weight mediator.
    \item \emph{Draft selection:} have a ranker (Llama 3.1 8B or Phi-4 14B) choose which of five drafts ``the widest range of these citizens'' would endorse \citep{bakker2022finetuning}.
    \item \emph{Draft merging:} ask the mediator to combine five drafts and retain only assignments found in at least three \citep{wang2023selfconsistency}.
\end{itemize}
Each set of drafts was ranked twice, in different orders. Selection and merging were tested on the three prompts asking for concrete proposals: concrete agreement, concrete differences and legislative compromise. %

For each task, we compared the outputs of two separate pools of five drafts with two pairs of randomly chosen drafts. Because safeguards change how often runs assign anyone, we counted differing pairs over all compared pairs, $D_{\text{all}}=\text{(pairs with differing assignments)}/\text{(all compared pairs)}$, so that pairs in which neither run assigns anyone are included and every method is compared on the same tasks. 
We classified a safeguard as useful if it reduced $D_{\text{all}}$ by at least 15 percentage points relative to random drafts (about a quarter of the 57--63\% found between random drafts; Appendix Table~\ref{tab:remedydenom}) and the 95\% confidence interval excluded zero. 
Because removing assignments also lowers $D_{\text{all}}$, we required merged statements to retain their level of detail and to have at least 80\% of assigned-party names supported by three or more source drafts (Appendix~\ref{app:controls}).

\emph{Objective (RQ3).} We compared agreement with differences separately within the plain and concrete styles. Within each style, only the objective clause changes. For each statement, the judge marked which of the thirty opinions it covered. On DeliberationBank, the judge also labeled each opinion as supporting or opposing the policy; an opinion is a \emph{minority position} if fewer than 40\% of the thirty sampled opinions share its stance, and a \emph{higher-support position} otherwise. Opinions with unclear stances are excluded. On Polis, comments that fewer than 40\% of voters agreed with are considered minority positions, while comments that more than 60\% agreed with are considered higher-support positions; the middle range is left out to avoid borderline cases. The main results also hold under other thresholds (Appendix~\ref{app:rq3}). These categories classify positions by support among sampled opinions (DeliberationBank) or voters (Polis); no demographic information is used. We ran this agreement-versus-differences comparison on both datasets and on DeliberationBank samples~2 and~3 (Appendix Table~\ref{tab:models}), and tested whether an explicit instruction to ``reflect the substance of every citizen's view'' closed the gap \citep{zhang2024fair}.

\subsection{Measurement and validation}
\label{sec:measures}

\emph{LLM judges.} The main judge, GPT-6.1 Sol (Sol), read each statement used in the analyses and listed who is assigned, to do what and on which terms. For two randomly chosen pairs of runs in each rerun set, it also labeled how the two statements related, using the five relations defined above. For coverage, it marked which of the thirty opinions a statement expressed, endorsed or explicitly acknowledged. To detect overstated coverage, the judge saw each statement with thirty-five numbered opinions in mixed order, the thirty real ones and five \emph{decoys} from other questions, and marked which of them the statement covered. A statement about health insurance, for example, should not cover an opinion about refugee admissions, so every decoy the judge marks is a false match. We used a judge's coverage readings only if it marked fewer than 5\% of all decoys as covered. \emph{Coverage} of a group is the share of its opinions that a statement covers, and the \emph{coverage gap} is the coverage of higher-support positions minus that of minority positions, in percentage points: a statement that covers 50\% of higher-support and 30\% of minority positions has a gap of 20 percentage points, and a value of zero means both groups are covered equally often. The second judge, GPT-5.6 Luna (Luna), repeated every measurement. Because Sol, Luna and the mediator Terra come from the same developer, and judges may favor outputs from their own model family \citep{panickssery2024llm}, a judge from another developer, Claude Haiku 5.5 (Haiku), repeated the readings of the annotated items and the coverage readings of the comparison of objectives (Appendix~\ref{app:additionaljudges}).

\emph{Human annotation.} Three annotators independently labeled a validation sample of 58 statements and 46 pairs of runs, plus 6 control pairs from different questions, which gives 330 judgments overall. For each statement they wrote down the assignment frame $(b,a,\theta)$: who is assigned, to do what and on which terms, and for each pair they chose one of the five relations. The pairs included 41 of the 2{,}280 pairs compared under unchanged opinions (1.8\%). The sample covered both datasets, all four main mediators, all types of reruns and all six prompts. It was stratified by a preliminary reading of Luna, so that it included assignments a judge may have missed as well as assignments a judge may have added. Annotators did not know which model, condition or judge label belonged to an item (Appendix~\ref{app:annotation}). The annotations tested whether the judges read assignments as people did and whether differences between runs were real. Agreement was measured with Cohen's $\kappa$, which corrects the share of matching labels for agreement expected by chance (1 means full agreement and 0 means random chance level); each annotator was compared with the label on which the other two agreed, and each judge with each pair of annotators \citep{cohen1960coefficient}; overall agreement among the three annotators was summarized with Krippendorff's $\alpha$.

\emph{Analysis.} Statements written from the same opinions are not independent, so the 95\% confidence intervals for the main estimates resample whole questions or Polis conversations (a cluster bootstrap); intervals for the human annotations resample annotated items \citep{field2007bootstrapping,cameron2008bootstrap}. When comparing the two objectives, we gave each question equal weight and left out each question in turn to check that no single question drove the result; results that also include the legislative compromise and irreducible conflict prompts are given in Appendix~\ref{app:rq3}. The same annotations were used to compare the two judges, so their agreement with the annotators (Section~\ref{sec:res-repro}) describes the judges on this sample and is not an independent test.

\section{Findings}
\label{sec:results}

\subsection{RQ1: The same opinions yield different assignments}
\label{sec:res-repro}

\begin{figure}[htbp!]
  \centering
  \includegraphics[width=\textwidth]{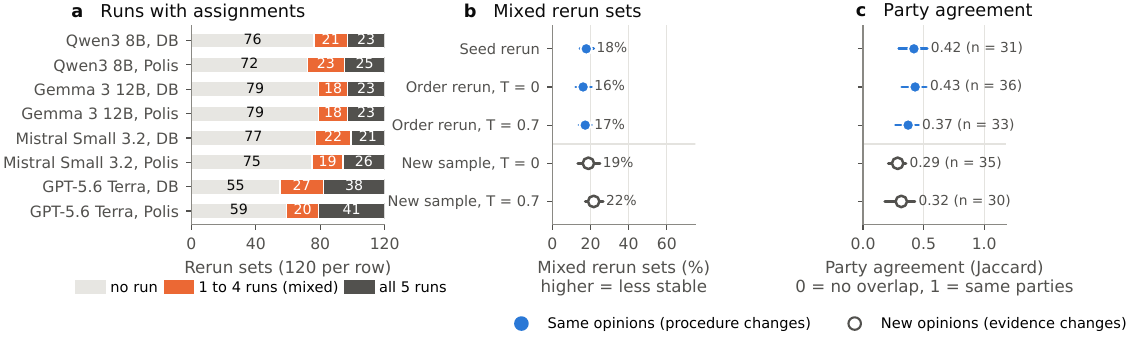}
  \Description{Left: one bar per mediator and dataset, split into rerun sets where no run, some runs or all five runs assign someone. Middle and right: for three open-weight mediators on DeliberationBank, the share of mixed sets and the party agreement for reruns on the same opinions and on new opinion samples.}
  \caption{\textbf{Reruns on the same opinions change whether and whom the proposals assign.} \textbf{a}, Each bar shows 120 rerun sets of one mediator on one dataset (repeated sampling at temperature 0.7 and order reruns at temperature zero). The bar is split by how many of the five runs assign someone: none (light), one to four (orange, mixed sets) or all five (dark). \textbf{b}, Share of mixed sets for the three open-weight mediators on DeliberationBank. Blue points: reruns on the same opinions with a new seed or a new order. White points: reruns on new samples of opinions to the same question, which serve as the reference. \textbf{c}, Party agreement (0, no overlap; 1, the same parties) for the rerun sets in which all five runs assign someone; $n$ is the number of such sets. Horizontal lines in b and c show 95\% confidence intervals for each level; differences between order and new-sample reruns, separately by temperature and pooled, appear with their own intervals in Appendix Table~\ref{tab:ordersample}.}
  \label{fig:stability}
\end{figure}

\emph{Whether anyone is assigned depends on the run.} We first examine whether a proposal assigns an action to an identifiable party. \figref{fig:stability}a shows how often the answer changes between runs on the same opinions. For every mediator and dataset, a substantial number of rerun sets is mixed: some of the five runs assign someone and others assign no one. Among the open-weight mediators, 16--18\% of rerun sets on DeliberationBank and 17\% on Polis were \emph{mixed} (16.9\% overall, 95\% CI 14.2--19.8); for Terra, 19.6\% were mixed (95\% CI 14.6--25.0). The DeliberationBank share varied by rerun type. Many sets contained no assignment in any run, particularly under the differences and irreducible conflict prompts. We therefore also examined sets where at least one run contained an assignment. Among these, 46.6\% of the 326 open-weight sets were mixed (95\% CI 39.2--55.0); the corresponding share for Terra was 37.3\% of 126 sets (95\% CI 28.8--45.8). Terra's lower share follows from assigning more often (Section~\ref{sec:res-controls}). In these mixed sets, publishing just one output hides the fact that another run could change whether anyone is assigned.

We checked whether inconsistent judging could explain the mixed sets by asking Sol to assess the same 1{,}027 statements again. Its label for whether anyone was assigned remained unchanged in 99.5\% of cases. Under the assumptions of our calculation, this level of inconsistency would produce at most 2.4\% mixed sets (Appendix~\ref{app:rq12}), well below the observed shares.

We also checked whether identical requests produced identical outputs at temperature zero. Within one session, Gemma 3 reproduced the same text in all 180 DeliberationBank repeats. Yet, when we reordered the opinions at temperature zero, every Gemma 3 set on DeliberationBank contained five different texts, and 15.0\% were mixed. %

\emph{Reordering opinions produces substantial changes in assignments.} We compared two ways of repeating a task: changing the order of the same opinions, and drawing a new sample of opinions for the same DeliberationBank question. With reordering, 16.7\% of five-run sets contained both statements with assignments and statements without. With new samples, this occurred in 20.3\% of sets (difference 3.6 percentage points, 95\% CI 1.1--6.1). New opinions changed whether anyone was assigned more often, but reordering the same opinions reached 82\% of that rate. These percentages are pooled over both temperatures; \figref{fig:stability}b shows them separately.

We also compared who received assignments, using only sets where all five runs contained assignments. Pooled over both temperatures, party agreement $R$ was 0.40 when the same opinions were reordered and 0.30 when new opinions were sampled (difference 0.10, 95\% CI 0.05--0.15). Higher values mean more overlap in who is assigned. When party names were also counted as matching if one appeared within the other (\emph{containment matching}; for example, ``city council'' and ``Bristol city council''), agreement was 0.49 for reordered opinions and 0.42 for new opinions. The gap between the two conditions narrowed (Appendix~\ref{app:rq12}). The qualifying sets differed between conditions because each had to contain assignments in all five runs. \figref{fig:stability}c shows the results separately by temperature. When temperatures are pooled and parties are matched by name, both measures show greater instability with new opinions than with reordering alone; the comparisons within each temperature are less conclusive (Appendix Table~\ref{tab:ordersample}).

\begin{figure}[!htb]
  \centering
  \includegraphics[width=\textwidth]{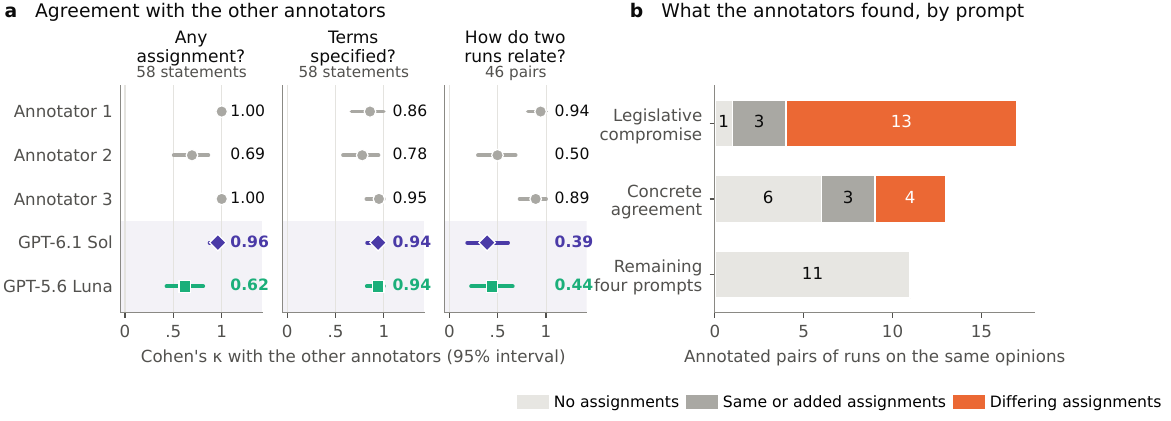}
    \Description{Left: three small panels, one per question, showing Cohen's kappa with intervals for the three annotators and the two LLM judges. Right: stacked bars of the annotated same-opinion pairs by prompt, showing differing assignments only under the legislative compromise and concrete agreement prompts.}
    \caption{\textbf{What the human annotators found.} \textbf{a}, Agreement with human reference labels (Cohen's $\kappa$; 1: full agreement; 0: agreement expected by chance), with 95\% bootstrap intervals. Each annotator is compared with the shared label of the other two, using only items on which those two agree. Each LLM judge is compared with all three two-annotator references, and its scores are averaged. ``Terms specified?'' asks whether an individual statement specifies at least one concrete term, such as an amount or deadline. Changes in terms between runs are assessed under ``How do two runs relate?''. \textbf{b}, The 41 annotated pairs generated from unchanged opinions, grouped by prompt and classified by the annotator majority; ``Remaining four prompts'' combines agreement, differences, concrete differences and irreducible conflict (Appendix Table~\ref{tab:humanprompts}).}
  \label{fig:human}
\end{figure}

\emph{Human readers confirm differences in assignments.} Automated comparisons may count two names for the same party, such as ``the city'' and ``the municipal government'', as different parties. We therefore asked three annotators to assess statements and pairs of runs, and compared their labels with those of the LLM judges (\figref{fig:human}a). On whether a statement contains an assignment and whether it specifies concrete terms, Sol agreed with the annotators about as well as they agreed with one another; Luna agreed less on whether an assignment is present.

Annotators agreed less consistently on whether two runs changed the parties or terms, imposed incompatible requirements, or simply added an assignment (Krippendorff's $\alpha=0.61$). Both LLM judges agreed less closely with the annotators on this task. Sol often labeled pairs as having differing assignments when the human majority found no assignment or only an additional one. We therefore use human labels to describe the types of changes between runs.

Of the 41 annotated pairs of runs on the same opinions, 23 contained an assignment in at least one run. In 17 of these 23 pairs, the majority of human annotators found differing assignments (\figref{fig:human}b). Using the detailed-label rule in Appendix~\ref{app:annotation}, these comprised 9 changes in terms, 7 changes in parties and 1 conflict. In the other 6, one run added an assignment to those of the other. Because the sample was stratified by preliminary judge labels, these counts establish that such differences occur and what types they take, rather than their overall frequency. Adjusting for the sampling design, we estimate that 22\% (95\% CI 14--31\%) of all 2,280 same-input pairs contained differing assignments according to the human-majority labels. This estimate assumes that pairs in the strata initially labeled as containing no assignments indeed contained none; sensitivity analyses are reported in Appendix~\ref{app:annotation}. The party-name extraction check in Appendix~\ref{app:annotation} assesses individual labels; it does not directly validate the rerun-level statistic $R$.

\emph{Assignments concentrate under agreement- and compromise-oriented prompts.} All annotated pairs with differing assignments came from the legislative compromise and concrete agreement prompts, which ask the mediator to specify mechanisms, amounts, eligibility, and/or deadlines. In the 11 pairs from the other four prompts, neither run contained an assignment. Across all rerun sets of the four mediators, mixed sets were most common under the agreement (34.7\%), concrete agreement (32.1\%) and legislative compromise (24.2\%) prompts, and rare under the two differences prompts (1.1\% each; Appendix Table~\ref{tab:promptspecificity}).

Requesting concrete detail increased the share of statements containing assignments from 9.4\% to 33.2\% on DeliberationBank and from 13.2\% to 40.5\% on Polis. This increase occurred mainly under agreement prompts. Among order reruns at temperature zero, concrete detail raised the share by 46.5 percentage points under agreement and 0.5 under differences on DeliberationBank (difference 46.0; 95\% CI 34.0--57.5). On Polis, it raised the share by 61.0 points under agreement and lowered it by 0.5 under differences (difference 61.5; 95\% CI 50.0--73.0). Prompts asking for actionable agreements or compromises therefore warrant particular attention in rerun audits.

\subsection{RQ2: Tested safeguards do not make assignments stable}
\label{sec:res-controls}

\begin{figure}[t]
  \centering
  \includegraphics[width=\textwidth]{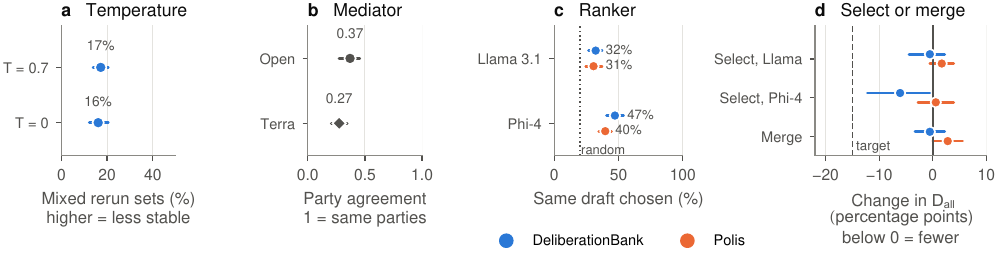}
  \Description{Four small interval plots for temperature, mediator, ranker consistency and the effect of selecting or merging drafts.}
  \caption{\textbf{None of the four safeguards made assignments stable.} \textbf{a}, Mixed rerun sets when only the order of the opinions changes, at temperature 0.7 and zero (180 DeliberationBank tasks, open-weight mediators). \textbf{b}, Party agreement, as in \figref{fig:stability}c, for the open-weight mediators and Terra (datasets pooled). \textbf{c}, How often a ranker chose the same draft when the five drafts were shown again in a new order (dotted line: 20\%, random choice). \textbf{d}, Change in $D_{\text{all}}$ (differing pairs over all compared pairs) after selecting or merging drafts, relative to two random drafts (below zero: fewer differences; dashed line: target of 15 percentage points). Horizontal lines show 95\% confidence intervals; the Terra comparison in b is tested in Appendix Table~\ref{tab:judges}.}
  \label{fig:controls}
\end{figure}

We tested four ways to reduce variation (\figref{fig:controls}). Panel a measures whether assignments appear in some runs and disappear in others; panel b measures agreement on who is assigned. Panel c checks whether reordering drafts changes the selected draft. Panel d measures whether selection or merging reduces the share of pairs with different parties, different terms or conflicting requirements.

\emph{Temperature zero.} Mixed sets remained almost as common: 16.1\% at temperature zero and 17.2\% at temperature 0.7 (\figref{fig:controls}a). The difference was $-$1.1 percentage points (95\% CI $-$5.0 to 2.8), providing no clear evidence of a reduction. In a separate check, repeating identical prompts at temperature zero in a later computing session changed most texts (Appendix~\ref{app:rq12}).

\emph{Commercial mediator.} Terra made assignments more often than the open-weight mediators: at least one of five runs contained an assignment in 53\% of its rerun sets, against 36\% for the open-weight mediators under the same rerun types. Among all its rerun sets, 19.6\% were mixed, against 16.8\% for the open-weight mediators under the same rerun types. Its lower share of 37.3\% among sets with an assignment (Section~\ref{sec:res-repro}) therefore comes from having more sets with an assignment, and does not show greater agreement between runs. We compared party agreement among sets where all five runs contained assignments. Agreement was 0.27 for Terra and 0.37 for the open-weight mediators (\figref{fig:controls}b). After adjustment for completeness and rerun type, the estimated difference was $-0.03$ on DeliberationBank and $-0.02$ on Polis; both confidence intervals included zero (Appendix Table~\ref{tab:judges}). These results provide no evidence that Terra was more stable.

\emph{Selecting a draft.} We asked a second model, the ranker, to choose one of five drafts. We then reordered the same drafts and asked it to choose again. It selected the same draft in 32--47\% of cases on DeliberationBank and 31--40\% on Polis, depending on the ranker (\figref{fig:controls}c). Independent random choices would match 20\% of the time. The rankers selected whichever draft appeared first in 34--64\% of calls (Appendix Table~\ref{tab:rankerintervals}). Selection therefore remained sensitive to presentation order.

\emph{Merging drafts.} We asked each mediator to combine its five drafts, retaining only assignments found in at least three. Name matching found that 61\% of the parties assigned in merged statements on DeliberationBank, and 55\% on Polis, appeared in at least three source drafts. We also found specific terms supported by only one draft. For example, a merged Mistral proposal financed health insurance ``by a 2\% income tax increase for households earning above the national median'', although this term appeared in only one source draft (Appendix~\ref{app:controls}).

\emph{Did selection or merging reduce differences?} We compared pairs of selected or merged outputs with randomly chosen draft pairs, measuring the share with differing assignments (\figref{fig:controls}d). Neither method reached our target of a reduction of 15 percentage points (Section~\ref{sec:design}). The largest reduction was 6.1 percentage points for Phi-4 selection on DeliberationBank (95\% CI 0.6--12.2); no upper limit exceeded 12.2 points, so any target above that value gives the same conclusion. The other estimates showed no clear reduction. These estimates rely on Sol's pair labels, which overstate differences between runs, so the size of any improvement remains sensitive to judging errors. The changes in draft selection after reordering and the single-source terms retained in merges provide separate evidence of limitations. Selected drafts contained assignments about as often as random drafts.

\subsection{RQ3: The objective changes whose positions are included}
\label{sec:res-representation}

RQ3 asks whether the mediator's objective changes which opinions a proposal covers (Section~\ref{sec:design}). Appendix~\ref{app:rq3} gives a worked example (\figref{fig:rq3example}) and the results for the two reference prompts. %

\begin{figure}[t]
  \centering
  \includegraphics[width=0.8\textwidth]{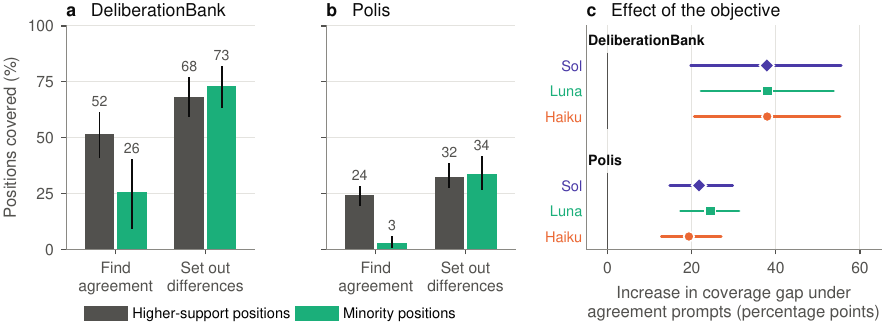}
  \Description{Left and middle: bars of higher-support and minority coverage under the two objectives for DeliberationBank and Polis. Right: interval plot of the objective contrast for the two datasets and three judges, labeled Sol, Luna and Haiku, all above zero.}
  \caption{\textbf{Coverage of positions under the two objectives.} \textbf{a, b}, Share of higher-support positions (dark bars) and minority positions (green bars) that a statement covers, under the prompts that ask the mediator to find agreement or to set out differences, as read by GPT-6.1 Sol; each panel pools 800 statements, and black lines show 95\% confidence intervals. \textbf{c}, Effect of the objective: how many percentage points larger the gap between higher-support and minority coverage is under the agreement prompts than under the differences prompts, for three LLM judges. Positive values mean a larger gap under the agreement prompts. Horizontal lines show 95\% confidence intervals.}
  \label{fig:objective}
\end{figure}

Under the agreement prompts, Sol judged minority positions to be covered less often than higher-support positions (\figref{fig:objective}a, b). Coverage was 26\% versus 52\% on DeliberationBank, and 3\% versus 24\% on Polis. Asking the mediator to set out differences increased coverage of both groups and brought their rates closer together: minority positions were covered in 73\% versus 68\% for higher-support positions on DeliberationBank, and 34\% versus 32\% on Polis. All three judges found a larger gap under agreement, although their estimates differed.

Compared with the differences prompts, the agreement prompts increased the coverage gap by 37.9 percentage points on DeliberationBank (95\% CI 19.7--55.5) and 21.8 points on Polis (95\% CI 14.8--29.7; \figref{fig:objective}c). We call this increase the \emph{objective contrast}. Panel c averages the results across questions that contain both kinds of position, so its estimates cannot be calculated directly from the pooled bars in panels a and b.

Luna and Haiku found similar contrasts (Appendix~\ref{app:additionaljudges}). The finding also held for two additional samples of thirty opinions per DeliberationBank question (samples~2 and~3), tested with a partly different set of mediator models. In each sample, the contrast remained positive when any one question was excluded (Appendix Table~\ref{tab:forest}).

To test whether repeated agreement statements recover omitted minority positions, we counted a position as covered if any of the five agreement statements for the same task covered it. Together, these statements covered 33.8\% of minority positions on DeliberationBank and 8.6\% on Polis. A single differences statement covered, on average, 72.1\% of minority positions on DeliberationBank and 32.1\% on Polis. It therefore covered more minority positions than the five agreement statements combined: 38.3 percentage points more on DeliberationBank (95\% CI 23.6--51.8) and 23.6 more on Polis (16.7--30.4; differences computed before rounding). Analyses using Luna and DeliberationBank sample~3 showed the same pattern (Appendix~\ref{app:rq3}).

Which positions appeared to be covered also varied between runs. When the opinions were reordered, Sol's assessment switched between covered and not covered for 18.8\% of positions on DeliberationBank and 17.1\% on Polis. When Sol assessed an unchanged statement again, the corresponding rates were 8.5\% and 3.1\%. The higher rates between reruns suggest that changes in the statements contribute to the variation, as does inconsistency in the judge.

The objective contrast persisted after adjusting for statement length, with and without the concreteness instruction, under alternative minority thresholds and with Luna's stance labels. On Polis, excluding 29 self-descriptions such as ``I live in a rural area'' increased it from 21.8 to 44.1 percentage points (Appendix~\ref{app:rq3}). Agreement statements that covered many positions still favored higher-support positions, so a statement can cover many positions while omitting those with less support. %

Sol and Haiku counted fewer than 0.5\% of decoy opinions from unrelated questions as covered on both datasets, so they rarely matched statements to unrelated content. However, the judges may count generic phrases as coverage (\figref{fig:rq3example}). How accurately they recognized paraphrases of relevant opinions is less certain (Appendix~\ref{app:additionaljudges}).

Finally, asking the mediator to ``reflect the substance of every citizen's view'' increased coverage of both groups. Higher-support positions gained more, so the coverage gap grew from 12.6 to 16.6 percentage points on DeliberationBank and from 10.4 to 22.4 on Polis (Appendix Table~\ref{tab:coverage}). On the odds scale, the interaction analysis supported a larger gain for higher-support positions on Polis; the result on DeliberationBank remained uncertain (Appendix~\ref{app:rq3}).

\section{Discussion}
\label{sec:discussion}

LLMs are increasingly used to summarize debates and to assign responsibilities to different parties. For such proposals to be legitimate and acceptable, the summaries and the assignments based on them must be transparent, and their variation across runs must be disclosed. Participants must also keep control over four steps: summarizing opinions, suggesting assignments, presenting them and committing to them (Table~\ref{tab:values}). Our results show deficits in the first three steps: summaries and assignments vary across runs and depend on the objective, and these choices are easily overlooked when organizers present a single output as the collective conclusion. A mediator should summarize opinions only when contributors have asked for it and agreed. Its objective should be disclosed, because the objective changes whose positions remain in the summary. Diverse positions are a precondition for collective intelligence \citep{page2007difference}, so a summary that drops minority positions narrows the options a group can consider. Assignments appeared mainly when the prompt asked for concrete detail, and some included amounts and deadlines that no opinion mentioned (\figref{fig:example}). Because assignments depend so much on the request, a mediator should suggest them only when participants ask for them and agree. Because they also vary across runs, a presented assignment should come with its alternatives, a record of how it was generated and selected, and the opinions that support it \citep{moreau2013prov,zaghouani2026accountable,mahajan2026participatory}. An assignment becomes a commitment only when participants explicitly endorse it and the assigned party accepts it. Until then, participants can contest the assignment, the objective and any claim of agreement \citep{alfrink2023contestable}. If any of these steps is skipped, the mediator could turn from a tool that serves participants into an instrument of power. It would then serve whoever controls its prompt and output selection, whether an organizer, a platform or a company.

\begin{table}[t]
\centering\footnotesize
\setlength{\tabcolsep}{4pt}
\caption{\textbf{Requirements for AI-supported participation, by process step.} Each row links a step and a value to a finding of this study and to a requirement that can be tested. The last row concerns the audit itself.}
\label{tab:values}
\begin{tabularx}{\textwidth}{@{}>{\raggedright\arraybackslash\itshape}p{1.35cm}>{\raggedright\arraybackslash\bfseries}p{2.0cm}YY@{}}\toprule
\textnormal{Step} & \textnormal{Value} & Finding & Requirement \\ \midrule
Summarize & Pluralism & Agreement prompts cover minority positions less than differences prompts (\figref{fig:objective}). & Summarize only with contributors' consent; state the objective, report coverage and retain unresolved disagreements. \\
Suggest & Human agency & Assignments appeared mainly when the prompt asked for concrete detail (9--13\% to 33--41\% of statements) and included terms no opinion mentioned (\figref{fig:example}). & Generate assignments only when participants request them. \\
Present & Transparency, equal treatment & The same opinions, run again or reordered, produce different assignments (Figs.~\ref{fig:example}, \ref{fig:stability}). & Show alternative assignments; test and disclose sensitivity to opinion order; record runs and the source of each assignment. \\
Present & Contestability & Reordering drafts changes selections; merges retain terms found in only one draft (Section~\ref{sec:res-controls}). & Open draft selection and merging to review and challenge. \\
Commit & Human agency & In 17 of 23 annotated pairs with an assignment, runs differed in parties or terms or conflicted (\figref{fig:human}b). & Treat assignments as suggestions until participants endorse them and the assigned parties accept them. \\
Audit & Reliable measurement & Judges report differences that human annotators do not (\figref{fig:human}a). & Validate audit results against human annotations. \\
\bottomrule\end{tabularx}
\end{table}

A practical audit generates five proposals from the same opinions, including reordered inputs, compares who is assigned, to do what and on which terms, and reports which assignments recur and which appear only in some runs. Five runs give an initial check; they do not establish that all alternatives have been found, and agreement between judges does not establish accuracy without human annotation. Presenting several drafts adds work for participants and creates room for selective presentation, so how to keep alternatives manageable should be tested with participants.

\emph{Limitations.}\label{sec:limitations}
The main experiments examine four mediators processing thirty opinions in English in a single pass, across two datasets with ten questions each. Larger groups, longer proposals and repeated interaction may change the findings and require adapted audits. Only DeliberationBank supports our comparison with five separate samples of opinions per question. Human annotators checked the assignment measures and found that the judges overstate differences between runs. They did not check coverage (RQ3). The three judges agreed on the direction of the objective contrast but differed in its size, and they may share errors in recognizing paraphrased opinions. Finally, we measure whether assignments remain stable across runs, but do not systematically assess whether each proposed assignment is justified by the source opinions or would be endorsed by participants. %

\section{Conclusion}
\label{sec:conclusion}

This study shows how turning citizens' opinions into a collective proposal can introduce choices that remain hidden when only one output is presented. The same opinions produced different assignments across runs, none of the four safeguards made them stable, and the request to find agreement gave higher-support positions a greater coverage advantage than the request to describe differences. Together, these findings show that a proposal reflects not only the citizens' input but also choices made during generation and the organizer's objective. Making those choices visible would allow participants to examine alternative assignments, question omissions and decide what they support. AI-mediated proposals should therefore remain open to deliberation, with their objectives disclosed and unresolved disagreements available for participants to consider.

\section*{Ethical considerations statement}
This study uses the public DeliberationBank (MIT license) and Polis openData (CC BY 4.0) datasets as published. Three informed annotators independently assessed model outputs and are reported anonymously. All generated statements are labeled as unendorsed synthetic proposals. While sensitivity to opinion order could be misused to steer outcomes, our proposed rerun-based audit can help detect such manipulation and serves as a check alongside broader requirements for democratic legitimacy.

\section*{Generative AI usage statement}
Anthropic's Claude (model Claude Opus 5.5) was used to assist with verification of bibliographic records,
and with debugging data-collection and analysis code. The language models used as mediators and judges in the study are part of the research design and are described in Section~\ref{sec:design}.

\bibliographystyle{unsrtnat}
\bibliography{references}

\appendix
\counterwithin{figure}{section}
\counterwithin{table}{section}

\section{Summary of related work}
\begin{table}[!htb]
\centering\small
\caption{\textbf{What earlier work evaluates and what this study adds.}}
\label{tab:literature}
\begin{tabularx}{\textwidth}{@{}p{3.4cm}YY@{}}\toprule
Study & Focus of evaluation & Variation considered \\ \midrule
\citet{bakker2022finetuning} & Approval of consensus statements & Candidates ranked by a reward model \\
\citet{tessler2024ai} & Endorsement and inclusion of minority critiques & Rounds of human critique \\
\citet{zhu2025can} & Representativeness of 18 summarizers & Input size; position of an opinion \\
\citet{fish2026generative}; \citet{low2026habermolt} & Representation; actionability & Selection among candidates; revision by participants \\
\citet{najera2026when}; \citet{majumdar2026generative} & Agreement between models; fairness of LLM votes & Different models; repeated votes \\
\citet{cooper2024arbitrariness} & Arbitrariness of classifier decisions & Retraining replicates \\
\midrule
\textbf{This study} & whether and whom a proposal assigns; coverage of minority positions & New seeds and new orders of unchanged opinions, against new opinions; safeguards; objectives \\
\bottomrule\end{tabularx}
\end{table}

\section{Data, prompts and study counts}
\label{app:data}\label{app:prompts}\label{app:ledger}

\begin{table}[t]
\centering\footnotesize
\caption{\textbf{Models and their roles.} Mediators write statements; rankers choose among existing drafts; judges read and compare statements.}
\label{tab:models}
\begin{tabularx}{\textwidth}{@{}p{4.6cm}Y@{}}\toprule
Models & Role \\ \midrule
Qwen3 8B; Gemma 3 12B; Mistral Small 3.2 24B & Main open-weight mediators for the primary analyses on both datasets; each also merges its own drafts (RQ2). \\
GPT-5.6 Terra & Main commercial mediator on both datasets. \\
Qwen3 8B; Gemma 3 12B; Llama 3.1 8B; Phi-4 14B & Mediators for DeliberationBank samples 2 and 3 (RQ3). \\
Llama 3.1 8B; Phi-4 14B & Rankers that select one draft from five (RQ2). \\
GPT-6.1 Sol; GPT-5.6 Luna & Main and secondary LLM judges for statements used in the analyses. \\
Claude Haiku 5.5 & Judge from another developer, used to check the main judges. \\
\bottomrule\end{tabularx}
\end{table}

\emph{Opinion sampling.} DeliberationBank contains 3,199 responses across ten questions. For each DeliberationBank question, we selected thirty opinions by taking every tenth response from the stored response list. Sample~1 begins with the first response, sample~2 with the sixth, and sample~3 with the fourth. These samples contain different opinions. The new-opinion comparison uses five groups beginning with the first, third, fifth, seventh and ninth responses; its first group contains the same opinions as sample~1.

Polis comments must have at least three words and ten votes and must not have been moderated out. We drew thirty comments using a pseudo-random seed derived from the SHA-256 hash of the conversation identifier. These comments remain fixed across seed and order reruns.
The Polis conversations are Bowling Green/Warren County, San Juan Islands land use, New Zealand biodiversity, Canadian electoral reform, affordable housing, tax policy, universal basic income, freshwater quality, Brexit and NFL concussions. They contain 1,497 eligible comments in total. Of the 300 sampled comments, 204 have agreement above 60\%, 50 below 40\%, and 46 fall between those cut-offs. The last conversation has no minority comment in its sampled thirty; minority coverage is therefore estimated from the remaining conversations. The source identifiers and sampled opinions are provided with the data.

\emph{Mediator prompts.} The four comparable prompts share the following stem and ending. The thirty opinions replace \texttt{\{opinions\}}, one per line and each preceded by a dash. The public question itself is not inserted into this prompt.
\begin{quote}\small
You are mediating a public policy debate. Below are the views of citizens on this issue.\par
\texttt{\{opinions\}}\par
Write a single statement of 2 to 4 sentences.\par
\emph{[One objective clause and one style clause from Table~\ref{tab:promptclauses}.]}\par
Return only the statement.
\end{quote}

\begin{table}[!htb]\centering\footnotesize
\caption{\textbf{Exact clauses in the four comparable mediator prompts.} We cross the two objectives with the two style clauses. The concrete clause replaces the plain-language clause. Comparisons between styles therefore measure the effect of this replacement.}
\label{tab:promptclauses}
\begin{tabularx}{\textwidth}{@{}p{2.5cm}Y@{}}\toprule
Clause & Text \\ \midrule
Find agreement & Your statement must set out what these citizens can agree on, in terms that people on opposing sides would both accept. \\
Set out differences & Your statement must set out where these citizens differ, as they actually differ. Do not look for agreement and do not propose a reconciliation. \\
Concrete & Be concrete: say who pays, how much, who qualifies, what happens if the rule is broken, and by when. \\
Plain language & Write in plain, direct language that a general newspaper reader could follow without any difficulty or specialist knowledge. \\
\bottomrule\end{tabularx}\end{table}

The first four prompts combine two objectives (find agreement or set out differences) with two styles (plain or concrete), using the clauses in Table~\ref{tab:promptclauses}. Within each style, the prompts differ only in the objective clause, so the two objectives can be compared while the remaining instructions stay fixed.

The two additional prompts ask the mediator to undertake broader tasks. The \emph{legislative compromise} prompt introduces a ``technocratic policy arbitrator'' and asks for a policy settlement with statutory mechanisms, fiscal allocations, conditional thresholds and enforcement procedures. The \emph{irreducible conflict} prompt introduces an ``agonistic pluralist mediator'' and asks it to identify unresolved conflicts and propose rules of engagement without forcing consensus. These prompts extend the audit of assignment stability to other approaches to mediation. Because they change the mediator's role, task and requested content together, they do not isolate the effect of the objective or of concreteness, so the main comparison of objectives excludes them; results that include them are reported in Appendix~\ref{app:rq3}.

All six full prompts and the judge, selection and merging prompts are included with the reproduction code. Generations use at most 512 output tokens, with reasoning disabled, and either temperature 0.7 with seeds 101--105 or temperature zero; Terra's interface does not take a seed, so its repetitions at temperature 0.7 are independent sampling calls. A prefatory phrase is removed when present; raw outputs are retained.

\emph{Counts and reuse.} Table~\ref{tab:ledger} counts each generation once, including identical texts produced in different calls. The same saved output may support several analyses. The core same-opinion comparison uses 5,700 statements in 1,140 five-run sets, with two sampled pairs per set (2,280 pairs). For DeliberationBank, the 900 open-model seed outputs include 600 Qwen3 and Gemma 3 outputs reused from the four-model grid and 300 separately generated Mistral outputs. The new-opinion reference adds 1,800 statements and 720 pairs. We pool the human annotations while retaining each sampling pool's original weights.

\begin{table}[!htb]\centering\footnotesize
\caption{\textbf{Generation ledger.} DB: DeliberationBank; $q$: questions or conversations; $m$: mediators; $p$: prompts. Counts refer to generated statements, including outputs not used in the reported analyses. Reused outputs are counted only in their generating row.}
\label{tab:ledger}
\begin{tabularx}{\textwidth}{@{}Yp{5.8cm}r@{}}\toprule
Component & Design arithmetic & Statements \\ \midrule
DB: order and new-opinion reruns & $10q\times3m\times6p\times4$ configurations $\times5$ & 3,600 \\
DB: additional Mistral seed outputs & $10q\times1m\times6p\times5$ & 300 \\
Polis: open-model seed and order & $10q\times3m\times6p\times2\times5$ & 1,800 \\
Terra: seed and order, both datasets & $2\times10q\times1m\times6p\times2\times5$ & 1,200 \\
DB: four-model grid, three samples & $3\times10q\times4m\times6p\times6$ & 4,320 \\
DB: Mistral identical-prompt repetitions & 180 saved responses & 180 \\
DB: other identical-prompt repetitions & $2m\times60$ prompts $\times3$ & 360 \\
Polis: later- and same-session repeats & $3m\times(300+60)$ & 1,080 \\
Selection candidate drafts, both datasets & $2\times10q\times3m\times3p\times10$ & 1,800 \\
Merged drafts, both datasets & $2\times10q\times3m\times3p\times2$ & 360 \\
Coverage instruction, four arms & $2\times10q\times3m\times2p\times5\times4$ & 2,400 \\
\midrule
Total & & 17{,}400 \\
\bottomrule\end{tabularx}\end{table}

The grid contains 720 temperature-zero outputs not used in the reported analyses. Samples~1 and~2 also support a comparison across random seeds. For the additional objective comparisons, sample~2 uses one seed per task, giving 240 statements across all six prompts; sample~3 uses five seeds per task, giving 1,200. Restricting these comparisons to the four agreement and differences prompts gives 160 statements for sample~2 and 800 for sample~3. For DeliberationBank sample~1 and Polis, the order reruns produce 1,200 statements per dataset across all six prompts and four main mediators, including Terra. The main objective comparison uses the 800 statements generated with the four agreement and differences prompts. Selection attempted 1,440 rankings; 1,437 were parsed (718 DB, 719 Polis), generating no new statement. The coverage-list control uses 600 additional Phi-4 opinion summaries, which are not collective proposals. The judges each assess 13,860 statements, 3,900 pairs, 6,621 coverage items and 900 DeliberationBank opinions; the coverage count includes 221 repeated readings and 160 planted-sentence checks. Repeated assignment readings add 1,027 Sol and 3,081 Luna assessments, and Sol's effort check adds 184 assessments. The additional Haiku checks add 1{,}854 assessments (58 statement readings, 196 pair readings and 1{,}600 coverage readings), counted separately below. Thus there are 26,492 Sol and 28,362 Luna assessments, separately from the generation count. The machine-readable ledger derives and checks these totals from saved records.

\section{Human annotation and rate estimation}
\label{app:annotation}

\emph{Instructions and sample.} Three annotators worked independently, without model identities, conditions or judge labels. They listed each statement's identifiable assigned parties, actions and specified terms. For pairs, they chose the relations defined in Section~\ref{sec:concept} and said whether an affected person would care which proposal was adopted. The human instructions provided worked examples and two additional rules: one-sided assignments usually count as an added assignment; otherwise overlapping categories follow conflict, different parties, different terms, added assignment, same assignments. The LLM prompts use the same definitions. We use the human majority as an interpretive reference and do not treat it as adjudicated ground truth.

\begin{table}[!htb]\centering\footnotesize\setlength{\tabcolsep}{4pt}
\caption{\textbf{The annotated sample.} Constrained stratified random sampling used dataset, sampling pool and original GPT-5.6 Luna labels, with mediator and task allocation constraints, so that missed and spurious assignments could both appear. Statements and compared pairs in the rerun data; statements, real pairs of runs and cross-topic control pairs annotated. Each item was labeled by all three annotators.}
\label{tab:annotsample}
\begin{tabular}{@{}>{\raggedright\arraybackslash}p{5.4cm}rrrrr@{}}
\toprule
Stratum & Statements & Pairs & Annotated statements & Annotated pairs & Control pairs \\
\midrule
DeliberationBank: shuffled order, new sample; GPT-5.6 Terra & 4,200 & 1,680 & 24 & 20 & 4 \\
DeliberationBank: random seed & 900 & 360 & 18 & 10 & 0 \\
Polis: random seed, shuffled order & 2,400 & 960 & 16 & 16 & 2 \\
\midrule
Total & 7,500 & 3,000 & 58 & 46 & 6 \\
\bottomrule\end{tabular}\end{table}

Table~\ref{tab:annotsample} gives the annotated sample by stratum. The annotated statements comprise 17 Qwen3, 17 Gemma 3, 13 Mistral and 11 Terra outputs; the real pairs comprise 11, 12, 14 and 9 respectively. All three people labeled every item: 58 statements, 46 real pairs and six controls, or 330 annotations. Controls are excluded from coverage, agreement and calibration. None of their 18 readings was ``same assignments''; the majority labeled four different parties, one added assignment and one with no assignment in either. Sol and Luna did not assess these six cross-question controls; the additional judges used separate identical-copy controls described below.

\emph{Agreement and disagreement.} We collapse pair labels into three classes for agreement and calibration: nobody assigned; same or added assignments; differing assignments. Annotators can agree that assignments differ without agreeing on the type of difference. Such pairs remain in the analysis. When no detailed label has a majority, we use the highest-priority label selected by an annotator, following the instruction order above. This gives the breakdown of 9 pairs with different terms, 7 with different parties and 1 with conflicting assignments among the 17 differing same-input pairs. Table~\ref{tab:poolagreement} reports each reader against the other annotators' agreed labels, with item-bootstrap intervals. Reference disagreements are excluded separately for each annotator pair, so retained items vary. Human Krippendorff's $\alpha$ is 0.79 for whether an assignment is present, 0.76 for whether concrete terms are specified and 0.61 for pair relations. Two annotators agree closely on pair relations, while the third differs more; majority agreement therefore should not be read as unanimity. Sol labels 15 of the 46 pairs as differing assignments when the majority does not, and detects all 20 majority-positive pairs. Luna overcalls eight and detects 17 of 20. Annotators also said whether an affected person would care which proposal was adopted. They answered yes for all 23 same-input pairs that the majority considered to propose assignments, unanimously for 20. The worked examples in the instructions link this answer closely to the relation label, so we do not treat it as independent evidence of consequence.

\begin{table}[!htb]\centering\footnotesize\setlength{\tabcolsep}{4pt}
\caption{\textbf{Agreement with human reference labels.} Each annotator is compared with the shared label of the other two, using only items on which those two agree. Each LLM judge is compared with all three two-annotator references, and its scores are averaged. Values are Cohen's $\kappa$, with 95\% intervals from 2,000 bootstrap resamples of annotated items. Pair relations use three classes: nobody assigned, same or added assignments, and differing assignments.}
\label{tab:annotators}\label{tab:poolagreement}
\begin{tabular}{@{}lccc@{}}
\toprule
Reader & Anyone assigned? (58) & Terms specified? (58) & How do two runs relate? (46) \\
\midrule
Annotator 1 & 1.00 [1.00, 1.00] & 0.86 [0.67, 1.00] & 0.94 [0.81, 1.00] \\
Annotator 2 & 0.69 [0.50, 0.86] & 0.78 [0.58, 0.94] & 0.50 [0.30, 0.69] \\
Annotator 3 & 1.00 [1.00, 1.00] & 0.95 [0.82, 1.00] & 0.89 [0.73, 1.00] \\
GPT-6.1 Sol & 0.96 [0.87, 1.00] & 0.94 [0.83, 1.00] & 0.39 [0.19, 0.61] \\
GPT-5.6 Luna & 0.62 [0.43, 0.81] & 0.94 [0.82, 1.00] & 0.44 [0.22, 0.66] \\
\bottomrule\end{tabular}\end{table}

For the separate party-name check, a label enters the human reference when at least two annotators list it. We lower-case and trim names for exact matching. Containment matching also counts two names as the same party when one appears within the other; each name can be matched to at most one name on the other side. Exact-match F1 is 78\% for Sol and 58\% for Luna. This is a lexical check of party labels and does not validate extracted actions or terms.

\emph{Weighting the combined annotations.} Sampling used original Luna labels within three pools, with constraints balancing tasks and mediators. Table~\ref{tab:strata} gives population and annotated counts. We retain the original sampling weights when combining the annotations. We estimate two rates for pairs generated from unchanged opinions. The first includes pairs in which the human majority identifies an assignment in at least one statement. The second includes pairs that Sol identifies as containing an assignment. Both use human labels to determine whether the assignments differ. Let $d_i$ indicate a seed or order pair, $b_i$ human-majority assignment presence, $m_i$ human-majority differing assignments, and $z_i$ Sol assignment presence. The indicator $d_i$ excludes new-opinion pairs from these rates. For stratum $h$, $N_h$ and $n_h$ are population and annotated counts. Define
\begin{equation}\label{eq:T}
\widehat T(x)=\sum_h\frac{N_h}{n_h}\sum_{i\in h}x_i .
\end{equation}
The human-defined rate and the rate within Sol-defined eligible pairs are respectively
\begin{equation}\label{eq:joint}
\widehat r_{\mathrm{human}}=\frac{\widehat T(d\,b\,m)}{\widehat T(d\,b)},
\end{equation}
\begin{equation}\label{eq:cond}
\widehat r_{\mathrm{Sol\ eligible}}=\frac{\widehat T(d\,z\,m)}{N_{d,z=1}},
\end{equation}
where $N_{d,z=1}$ is the known number of Sol-eligible same-input pairs in the full population. The rates can differ because humans and Sol do not identify the same pairs as containing assignments. Intervals use 4,000 within-stratum resamples. Sparse and singleton strata, task constraints and reuse of annotations to select the judge limit their interpretation.

\begin{table}[!htb]\centering\footnotesize
\caption{\textbf{Sampling strata of the human calibration.} Each stratum is defined by the label that GPT-5.6 Luna gave to every compared pair when the sample was drawn. Compared: pairs in the rerun data; annotated: pairs labeled by the three annotators, of which same input: seed or order pairs; weight: compared divided by annotated.}
\label{tab:strata}
\begin{tabularx}{\textwidth}{@{}>{\raggedright\arraybackslash}p{3.2cm}Yrrrr@{}}\toprule
Pool & Stratum (GPT-5.6 Luna label) & Compared & Annotated & Same input & Weight \\ \midrule
DB: order, new sample and Terra & different parties & 7 & 1 & 1 & 7.0 \\
 & different terms & 358 & 6 & 5 & 59.7 \\
 & conflicting assignments & 63 & 3 & 1 & 21.0 \\
 & nobody assigned & 1{,}086 & 3 & 1 & 362.0 \\
 & added assignment & 104 & 3 & 3 & 34.7 \\
 & same assignments & 62 & 4 & 4 & 15.5 \\
DB: open-model seed reruns & nobody assigned & 248 & 2 & 2 & 124.0 \\
 & at least one run assigns someone & 112 & 8 & 8 & 14.0 \\
Polis: seed and order & nobody assigned & 578 & 3 & 3 & 192.7 \\
 & at least one run assigns someone & 382 & 13 & 13 & 29.4 \\
\bottomrule\end{tabularx}\end{table}

\begin{table}[!htb]\centering\footnotesize
\caption{\textbf{Rate estimates using each annotator separately.} Human-defined same-input rate with the original sampling weights and one annotator's labels in place of the majority (\%; 95\% within-stratum bootstrap interval).}
\label{tab:calibrationreaders}
\begin{tabular}{@{}lrr@{}}\toprule
Reference & Estimate & Interval \\ \midrule
Annotator 1 & 69 & 47--89 \\
Annotator 2 & 62 & 37--80 \\
Annotator 3 & 77 & 57--98 \\
Majority & 71 & 49--91 \\
\bottomrule\end{tabular}\end{table}

\emph{Sensitivity of the weighted estimate.} Most compared pairs fall in strata that Luna labeled as containing no assignments, but few of these pairs were annotated. The human majority also found no assignments in those annotated pairs. Resampling these observations cannot reveal assignments that may have been missed in the unannotated pairs. Changing the label of the single annotated pair in the largest of these strata to a pair whose assignments do not differ gives 47\%; changing it to a pair whose assignments differ gives 80\%. If 5\% or 10\% of the Luna ``neither'' pairs in fact proposed assignments that do not differ, the estimate would be 64\% or 59\%. Post-stratifying by prompt instead gives 72\%. Replacing the majority with a single annotator's labels gives 69\%, 62\% and 77\% (Table~\ref{tab:calibrationreaders}). The unweighted count, 17 of 23 pairs with an assignment, does not depend on these assumptions; it describes the annotated sample and is not a population rate. Table~\ref{tab:humanprompts} gives this count by prompt.

Across all 2{,}280 same-input pairs, including pairs that assign nobody, the human-adjusted share with differing assignments is 22\% (14--31\%). This estimate assumes, as the annotated pairs in them indicate, that pairs in the ``neither assigns'' strata assign nobody; the sensitivity checks above show how the conditional rate changes otherwise. For the rate among human-defined assignment-containing pairs, restricting the analysis to configurations shared by both datasets gives 71\% (95\% CI 45--96\%; 36 annotated pairs). Deleting one annotated same-input pair at a time gives 67--77\% over 40 defined deletions; the remaining deletion removes its stratum's only annotation and has no defined estimate without another assumption. If the conditional rate uses an estimated rather than known Sol-eligible denominator, it is 52\% (34--69\%). These checks do not remove the uncertainty due to sampling.

\begin{table}[!htb]\centering\footnotesize
\caption{\textbf{The human evidence by prompt.} Annotated same-input pairs only. The majority determines whether at least one output proposes an assignment and whether assignments differ. These are observed sample counts, not population rates; sampling was stratified by the preliminary Luna labels and pool, not by prompt.}
\label{tab:humanprompts}
\begin{tabular}{@{}lrrr@{}}\toprule
Prompt & Annotated pairs & Propose an assignment & Differing assignments \\ \midrule
Agreement & 1 & 0 & 0 \\
Agreement + concrete & 13 & 7 & 4 \\
Differences & 2 & 0 & 0 \\
Differences + concrete & 3 & 0 & 0 \\
Legislative compromise & 17 & 16 & 13 \\
Irreducible conflict & 5 & 0 & 0 \\
\bottomrule\end{tabular}\end{table}

\section{Stability: supporting measurements}
\label{app:rq12}\label{app:judges}

\emph{Measurement noise and runtime nondeterminism.} Sol read 1{,}027 statements twice. It agreed on whether anyone is assigned in 99.5\% of cases ($\kappa$ 0.99) and gave party Jaccard 0.94 among statements assigning someone in both readings. Using the observed assignment prevalence and conditional disagreement rates, our calculation estimates that reading inconsistency alone would produce at most 2.4\% mixed sets under its assumptions. Repeated readings measure consistency; errors shared by both readings remain undetected. 

Gemma 3 reproduced 180 of 180 identical prompts exactly within a session, while all of its DeliberationBank order rerun sets at temperature zero contained five distinct texts (15.0\% mixed; party agreement 0.35 in the few sets in which all five runs assign someone). In a later session, the same prompts at temperature zero produced different text in 79.0\% of DeliberationBank and 85.0\% of Polis comparisons with the original order reruns, and assignment presence changed in 3.7\% of them. In a separate Polis repetition, agreement between sessions depended on which draw was compared, for reasons we could not identify; we therefore do not interpret exact-reproduction rates further.

\emph{Procedure compared with evidence.} Table~\ref{tab:ordersample} separates the two temperatures. At temperature zero, the interval for the ratio of mixed sets reaches 1; at temperature 0.7, the interval for the ratio of party disagreement includes 1. Pooled over both temperatures, both intervals exclude 1. Seed reruns have no new-sample counterpart and are not part of this comparison.

\begin{table}[!htb]\centering\footnotesize\setlength{\tabcolsep}{3.5pt}
\caption{\textbf{Reordering the same opinions compared with consulting new opinions.} DeliberationBank, three open-weight mediators, 180 matched tasks per temperature (10 questions $\times$ 3 mediators $\times$ 6 prompts). Mixed: five-run sets in which some but not all runs assign someone. Party agreement: Eq.~\ref{eq:R}, among $n$ sets in which all five runs assign someone. Ratios divide order-rerun instability by new-sample instability (mixed share; $1-R$). 95\% intervals resample questions.}
\label{tab:ordersample}
\begin{tabular}{@{}lrrrrrr@{}}\toprule
Temperature & Mixed, order (\%) & Mixed, new sample (\%) & Ratio, mixed & $R$, order ($n$) & $R$, new sample ($n$) & Ratio, $1-R$ \\ \midrule
0 & 16.1 & 18.9 & 0.85 (0.76--1.00) & 0.43 (36) & 0.29 (35) & 0.80 (0.73--0.88) \\
0.7 & 17.2 & 21.7 & 0.79 (0.68--0.91) & 0.37 (33) & 0.32 (30) & 0.92 (0.79--1.08) \\
Both & 16.7 & 20.3 & 0.82 (0.74--0.93) & 0.40 (69) & 0.30 (65) & 0.85 (0.79--0.93) \\
\bottomrule\end{tabular}\end{table}

Tables~\ref{tab:step} and \ref{tab:materiality} separate rerun configurations and keep the new-opinion reference visible. Table~\ref{tab:promptspecificity} shows how assignment presence depends on the prompt; pair eligibility and extracted assignment presence are separate judge decisions. Leaving out one question at a time in the same-input analysis gives party agreement from 0.27 to 0.43. Allowing containment matching gives 0.39--0.52 for the open mediators across all configurations, including the new-opinion reference; for the comparison of reordering with new opinions it gives 0.49 versus 0.42 (difference 0.08, 95\% CI 0.00--0.15), and for Terra versus the open mediators 0.41 versus 0.46 (difference $-$0.05, $-$0.13 to 0.03).

\begin{table}[!htb]
\centering
\footnotesize
\caption{\textbf{Party agreement by rerun configuration.} Open mediators are pooled unless Terra is named. None assign and some assign: mean Jaccard similarity (Eq.~\ref{eq:R}, with $J(\emptyset,\emptyset)=1$) and, in parentheses, the number of five-run sets in which no output or one to four outputs assign someone. Last column: mean Jaccard similarity when all five assign someone, question-bootstrap interval and number of qualifying sets. New-opinion samples change the evidence and serve as a reference.}
\label{tab:step}
\resizebox{\ifdim\width>\textwidth\textwidth\else\width\fi}{!}{%
\begin{tabular}{@{}llrrl@{}}
\toprule
Data & Configuration & None assign & Some assign & All five: Jaccard [95\% CI] ($n$) \\
\midrule
DB & Seed ($T{=}$0.7) & 1.00 (117) & 0.29 (32) & 0.42 [0.30, 0.53] (31) \\
DB & Order ($T{=}$0) & 1.00 (115) & 0.37 (29) & 0.43 [0.32, 0.52] (36) \\
DB & Order ($T{=}$0.7) & 1.00 (116) & 0.35 (31) & 0.37 [0.27, 0.46] (33) \\
DB & Resample ($T{=}$0) & 1.00 (111) & 0.36 (34) & 0.29 [0.21, 0.36] (35) \\
DB & Resample ($T{=}$0.7) & 1.00 (111) & 0.33 (39) & 0.32 [0.18, 0.42] (30) \\
DB, GPT-5.6 Terra & Seed ($T{=}$0.7) & 1.00 (27) & 0.45 (12) & 0.27 [0.16, 0.37] (21) \\
DB, GPT-5.6 Terra & Order ($T{=}$0) & 1.00 (28) & 0.44 (15) & 0.33 [0.19, 0.47] (17) \\
Polis & Seed ($T{=}$0.7) & 1.00 (113) & 0.32 (30) & 0.33 [0.18, 0.48] (37) \\
Polis & Order ($T{=}$0) & 1.00 (113) & 0.32 (30) & 0.31 [0.17, 0.47] (37) \\
Polis, GPT-5.6 Terra & Seed ($T{=}$0.7) & 1.00 (30) & 0.32 (9) & 0.29 [0.18, 0.39] (21) \\
Polis, GPT-5.6 Terra & Order ($T{=}$0) & 1.00 (29) & 0.38 (11) & 0.22 [0.16, 0.30] (20) \\
\bottomrule
\end{tabular}}
\end{table}

\begin{table}[!htb]
\centering
\footnotesize
\caption{\textbf{Differences between reruns, as labeled by the LLM judge.} Pairs: two random pairs per task. Eligible: pairs in which at least one statement assigns someone. Among eligible pairs: any change (added assignment, different parties, different terms or conflicting assignments); differing assignments (different parties, different terms or conflicting assignments; 95\% cluster-bootstrap CI); conflicting assignments. Last column: differing assignments as a share of all pairs.}
\label{tab:materiality}
\resizebox{\ifdim\width>\textwidth\textwidth\else\width\fi}{!}{%
\begin{tabular}{@{}llrrrlrr@{}}
\toprule
Data & Rerun type & Pairs & Eligible & Any change (\%) & Differing (\%) & Conflicting (\%) & Differing, all (\%) \\
\midrule
DB & Seed ($T{=}$0.7) & 360 & 148 & 97 & 78 [71, 84] & 5 & 32 \\
DB & Order ($T{=}$0) & 360 & 135 & 98 & 83 [76, 89] & 7 & 31 \\
DB & Order ($T{=}$0.7) & 360 & 142 & 98 & 77 [72, 84] & 6 & 30 \\
DB & Resample ($T{=}$0) & 360 & 141 & 97 & 80 [73, 88] & 8 & 31 \\
DB & Resample ($T{=}$0.7) & 360 & 143 & 96 & 79 [74, 83] & 8 & 31 \\
DB, GPT-5.6 Terra & Seed ($T{=}$0.7) & 120 & 80 & 98 & 88 [79, 96] & 1 & 58 \\
DB, GPT-5.6 Terra & Order ($T{=}$0) & 120 & 77 & 95 & 86 [78, 94] & 3 & 55 \\
Polis & Seed ($T{=}$0.7) & 360 & 145 & 99 & 88 [81, 94] & 5 & 36 \\
Polis & Order ($T{=}$0) & 360 & 147 & 96 & 83 [78, 89] & 11 & 34 \\
Polis, GPT-5.6 Terra & Seed ($T{=}$0.7) & 120 & 74 & 99 & 92 [88, 96] & 4 & 57 \\
Polis, GPT-5.6 Terra & Order ($T{=}$0) & 120 & 79 & 95 & 90 [82, 97] & 4 & 59 \\
\bottomrule
\end{tabular}}
\end{table}

\begin{table}[!htb]\centering\footnotesize
\caption{\textbf{Assignments under each prompt, with the opinions unchanged.} All four mediators, seed and order reruns. DeliberationBank additionally includes temperature-0.7 order reruns for the three open mediators. Outputs: generated statements; Assigned: percentage whose extracted frame names a party and an action; sets: groups of five outputs; mixed: sets in which only some outputs assign someone. Pairs: two sampled comparisons per set; eligible: pairs that Sol labels as proposing at least one assignment; differing: eligible pairs labeled different parties, different terms or conflicting assignments. Extraction and pair eligibility are separate judge readings; the pair judge counts many irreducible-conflict pairs as proposing assignments although few of these statements assign anyone, one reason we rely on human labels for pair relations. Counts are descriptive and have no human correction. The first four prompts form the factorial design; the final two are the legislative compromise and irreducible conflict prompts.}
\label{tab:promptspecificity}
\begin{tabular}{@{}llrrrrrrr@{}}\toprule
Data & Prompt & Outputs & Assigned (\%) & Sets & Mixed & Pairs & Eligible & Differing \\ \midrule
DB & Agreement & 550 & 18.7 & 110 & 40 & 220 & 46 & 21 \\
DB & Agreement + concrete & 550 & 65.8 & 110 & 37 & 220 & 181 & 162 \\
DB & Differences & 550 & 0.0 & 110 & 0 & 220 & 0 & 0 \\
DB & Differences + concrete & 550 & 0.5 & 110 & 2 & 220 & 12 & 9 \\
DB & Legislative compromise & 550 & 91.1 & 110 & 27 & 220 & 220 & 220 \\
DB & Irreducible conflict & 550 & 3.5 & 110 & 13 & 220 & 123 & 60 \\
Polis & Agreement & 400 & 25.2 & 80 & 26 & 160 & 36 & 26 \\
Polis & Agreement + concrete & 400 & 81.0 & 80 & 24 & 160 & 157 & 152 \\
Polis & Differences & 400 & 1.2 & 80 & 2 & 160 & 0 & 0 \\
Polis & Differences + concrete & 400 & 0.0 & 80 & 0 & 160 & 8 & 4 \\
Polis & Legislative compromise & 400 & 87.5 & 80 & 19 & 160 & 160 & 160 \\
Polis & Irreducible conflict & 400 & 3.5 & 80 & 9 & 160 & 84 & 47 \\
\bottomrule\end{tabular}\end{table}

\emph{Judge checks.} Frame completeness is 0 when no required action is identified, 1 for an action without an identifiable party, 2 for a party and an action, and 3 when terms are also specified. Both judges assess the same items with the same prompts. Their eligibility decisions differ, so Table~\ref{tab:reviewjudges} restricts the comparison to sets where both find assignments in all five outputs. Low party agreement remains. Repeated readings of identical statements are more consistent: Sol's party Jaccard is 0.97 at completeness 2 and 0.93 at completeness 3 (1{,}027 repeated readings), compared with 0.48 and 0.33 between generated outputs. Luna gives 0.83 and 0.88 on identical statements (3{,}081 readings), versus 0.46 and 0.26 between outputs. These checks assess the consistency of repeated readings across different eligible records; human agreement is assessed separately through the annotations. On the 24 statements and 20 pairs of the first sampling pool, Sol's medium reasoning setting gave the same assignment-presence agreement as low effort (100\%); three-class pair agreement was 60\% versus 55\%, with leave-one-out $\kappa$ 0.37 versus 0.34. We retained low effort; the sample was not held out.

\begin{table}[!htb]\centering\footnotesize
\caption{\textbf{Both judges on the same qualifying groups.} Groups contain five outputs, each assigning someone according to both judges. Mean actor-label Jaccard similarity (95\% question-cluster bootstrap interval). Counts describe groups, not independent questions.}
\label{tab:reviewjudges}
\begin{tabular}{@{}lllrrr@{}}\toprule
Data & Mediators & Configuration & Groups & Sol & Luna \\ \midrule
DB & Three open & Order (T=0) & 36 & 0.43 (0.31--0.52) & 0.38 (0.27--0.48) \\
DB & Three open & Order (T=0.7) & 32 & 0.37 (0.26--0.46) & 0.36 (0.24--0.47) \\
DB & Three open & Seed (T=0.7) & 30 & 0.42 (0.29--0.53) & 0.39 (0.25--0.49) \\
DB & Terra & Order (T=0) & 17 & 0.33 (0.19--0.47) & 0.26 (0.13--0.41) \\
DB & Terra & Seed (T=0.7) & 21 & 0.27 (0.16--0.37) & 0.27 (0.16--0.38) \\
Polis & Three open & Order (T=0) & 35 & 0.32 (0.18--0.48) & 0.30 (0.16--0.46) \\
Polis & Three open & Seed (T=0.7) & 34 & 0.30 (0.17--0.42) & 0.26 (0.17--0.34) \\
Polis & Terra & Order (T=0) & 20 & 0.22 (0.15--0.30) & 0.22 (0.14--0.32) \\
Polis & Terra & Seed (T=0.7) & 20 & 0.26 (0.16--0.35) & 0.23 (0.17--0.29) \\
\bottomrule\end{tabular}\end{table}

\subsection{Checks with a judge from another developer}
\label{app:additionaljudges}

We check the main judges with Anthropic's Claude Haiku 5.5 (adaptive thinking, medium effort). Haiku read the 58 annotated statements, 46 annotated real pairs, 46 repeated pair readings, 46 reversed pairs and 58 identical-copy pairs, with the same prompts as the main judges, and all 1,600 statements of the comparison of objectives (800 per dataset) for coverage. Human labels and other judges' relation or coverage decisions do not appear in the prompts; the coverage items reuse Sol's position summaries and groups, as described next. These are additional readings of existing annotation items and do not form a held-out human validation. Sol remains the main judge; the human-adjusted rates are unchanged.

For coverage, Sol's position texts, group labels and candidate order are held fixed, so this tests only the coverage reading and does not reconstruct the pipeline independently. Each assessment mixes thirty real positions and five decoys from other consultations. Candidate numbers identify positions; they are not ratings. Decoy selection divides selected decoys by all decoy opportunities. Low selection checks whether the judge rejects unrelated content; it cannot establish recall on real positions, and occasional overlap between consultations is possible.

Table~\ref{tab:additionaljudges} separates the human comparisons from these controls. Haiku finds differing assignments in all 20 pairs that the human majority labels as differing, but also in 12 of the other 26 pairs; Sol does so in 15. Thus a different developer does not remove the need for human adjustment. Haiku finds no differing assignments in any of the 58 identical-copy pairs. Its three-category labels agree on 43/46 same-order repeats and 43/46 reversed pairs. Haiku and Sol select fewer than 0.5\% of decoys on both datasets, below the 5\% screening criterion. These findings concern the tested configurations and do not rank the model families in general.

\begin{table}[!htb]\centering\footnotesize
\caption{\textbf{Additional-judge checks on fixed items.} The first two columns count decisions that match the annotators' majority: whether a statement assigns anyone (58 statements) and whether two runs have differing assignments (46 pairs); they describe this sample, not population accuracy. Decoys are positions from other consultations selected as covered, out of all decoy opportunities (five per statement read). A dash means Luna was not assessed on these decoys. Lower decoy selection is preferable. These are different tasks and denominators; no overall score is calculated.}
\label{tab:additionaljudges}
\begin{tabularx}{\textwidth}{@{}Yrrrr@{}}\toprule
Judge & Anyone assigned & Differing assignments & DB decoys & Polis decoys \\ \midrule
Sol & 57/58 & 31/46 & 1/4,000 & 4/4,000 \\
Luna & 47/58 & 35/46 & --- & --- \\
Haiku & 57/58 & 34/46 & 0/4,000 & 17/4,000 \\
\bottomrule\end{tabularx}\end{table}

The objective contrast recurs with Haiku on both datasets (Table~\ref{tab:forest}). Absolute coverage remains judge-sensitive: on DeliberationBank, Haiku's gaps are 16.8 percentage points under the agreement prompts and $-$21.2 under the differences prompts, against 34.5 and $-$3.4 for Sol (Table~\ref{tab:forest}). We treat this as evidence for an objective-dependent change in relative coverage, without assuming parity under difference-setting across judges.

\emph{Sensitivity to questions, group definitions and judge.} In the full Sol and Luna coverage analyses, the agreement-minus-differences gap remains positive after weighting eligible questions equally and after leaving out any one question. The direction also persists with minority cut-offs of 30\%, 40\% and 50\%, and with a strict-majority comparison group; these schemes change the groups and sometimes the eligible questions. Over all 180 matched tasks, temperature zero changes party agreement little: Luna gives a mean Jaccard change of 0.010 (95\% interval $-$0.012 to 0.036), while Sol gives 0.012 ($-$0.009 to 0.033). Both intervals include zero. Complete sensitivity outputs accompany the reproduction code.

\begin{table}[!htb]\centering\footnotesize
\caption{\textbf{Commercial comparison with each judge.} Difference in task-level party Jaccard after adjustment for mean frame completeness and rerun type (95\% question-bootstrap intervals). The comparison pools the tested configurations and is descriptive; it does not isolate model size, developer or reasoning capability.}
\label{tab:judges}
\begin{tabularx}{\textwidth}{@{}Yll@{}}\toprule
Terra minus open mediators & Sol & Luna \\ \midrule
DeliberationBank & \textminus 0.029 [\textminus 0.068, 0.010] & \textminus 0.032 [\textminus 0.090, 0.023] \\
Polis & \textminus 0.019 [\textminus 0.067, 0.038] & \textminus 0.031 [\textminus 0.086, 0.029] \\
\bottomrule\end{tabularx}\end{table}

\section{Controls: draft processing}
\label{app:controls}

We compare pairs of selected drafts and pairs of merged statements with randomly chosen draft pairs. The reduction criterion uses all compared pairs, including those containing no assignments; Table~\ref{tab:remedydenom} also reports results restricted to pairs containing assignments. Success requires a reduction of at least 15 percentage points and a confidence interval excluding zero. For merging, we additionally require mean completeness to remain within 0.5 of the source drafts and at least 80\% of party names to have support in three or more source drafts. Only 61\% of party names in DB merged outputs and 55\% in Polis occur in at least three drafts; 91\% and 95\% occur in any draft. These lexical matches concern assigned parties, not whether each proposed action or term is supported. Names with no match may reflect new assignments or different naming.

\emph{Terms kept from a single draft.} We searched merged statements for specific amounts and checked, after normalizing numbers, how many of the five source drafts contain them. Three cases illustrate the pattern. A Mistral Small 3.2 merge on health insurance (legislative compromise prompt) finances the program ``by a 2\% income tax increase for households earning above the national median''. A Qwen3 merge on refugees (legislative compromise) establishes a fund ``allocating \$500 million annually for housing, education, and employment support''. A Qwen3 merge on product obsolescence (concrete agreement) adds ``fines of up to \$500,000 per violation''. In each case, exactly one of the five drafts contains the amount, in the same context, although the merging instruction asks to keep only assignments found in at least three drafts.

\begin{table}[!htb]
\centering
\footnotesize
\caption{\textbf{Selection and aggregation: the two denominators.} The target of 15 percentage points uses the share of pairs with differing assignments among \emph{all} compared pairs, including pairs in which neither statement assigns anyone (column 4). Eligible pairs contain an assignment in at least one statement; the last column divides differing pairs by eligible pairs. Labels are those of the LLM judge.}
\label{tab:remedydenom}
\resizebox{\ifdim\width>\textwidth\textwidth\else\width\fi}{!}{%
\begin{tabular}{@{}llrrrr@{}}
\toprule
Data & Comparison & All pairs & Differing/all (\%) & Eligible pairs & Differing/eligible (\%) \\
\midrule
DB & Two random candidates & 180 & 57 & 106 & 97 \\
DB & Winners, Llama 3.1 ranks & 90 & 57 & 52 & 98 \\
DB & Winners, Phi-4 ranks & 90 & 51 & 52 & 88 \\
DB & Two merged statements & 90 & 57 & 55 & 93 \\
Polis & Two random candidates & 180 & 63 & 117 & 97 \\
Polis & Winners, Llama 3.1 ranks & 90 & 64 & 58 & 100 \\
Polis & Winners, Phi-4 ranks & 90 & 63 & 59 & 97 \\
Polis & Two merged statements & 90 & 66 & 59 & 100 \\
\bottomrule
\end{tabular}}
\end{table}

\begin{table}[!htb]\centering\footnotesize
\caption{\textbf{Denominators and uncertainty for ranker behavior.} First-position frequency uses parsed ranking calls; consistency uses pools with both orders parsed. Intervals resample questions (\%; 95\% CI).}
\label{tab:rankerintervals}
\begin{tabular}{@{}llrcrc@{}}\toprule
Data & Ranker & Parsed calls & First position & Paired pools & Consistency \\ \midrule
DB & Llama 3.1 8B & 360 & 51 (46--56) & 180 & 32 (27--37) \\
DB & Phi-4 14B & 358 & 34 (29--40) & 178 & 47 (41--54) \\
Polis & Llama 3.1 8B & 360 & 57 (55--60) & 180 & 31 (24--37) \\
Polis & Phi-4 14B & 359 & 64 (58--70) & 179 & 40 (35--45) \\
\bottomrule\end{tabular}\end{table}

\section{Objective and position coverage}
\label{app:rq3}

\begin{figure}[h!]
\centering\small
\setlength{\fboxsep}{6pt}
\fbox{\begin{minipage}{0.96\textwidth}
\textbf{One of the thirty opinions on public healthcare} \hfill \emph{opposing stance: 5 of 30 opinions}\par\smallskip
``I'd say that it highly depends on various factors. For one thing, you should have to be a citizen of a country to receive its healthcare. So in general, I'd say no.''
\end{minipage}}\par\medskip
\fbox{\begin{minipage}[t]{0.465\textwidth}
\textbf{Agreement prompt} \hfill \emph{not covered (Sol)}\par\smallskip
A majority of citizens believe access to basic healthcare is a fundamental need and that the government has a role to play in ensuring it. Many support government-backed health insurance to reduce financial burdens and improve public health outcomes, though concerns exist about cost, efficiency, and potential impacts on quality of care. Ultimately, there's a widespread desire to ensure more people can receive medical attention regardless of their income or employment status.
\end{minipage}}\hfill
\fbox{\begin{minipage}[t]{0.465\textwidth}
\textbf{Differences prompt} \hfill \emph{covered (Sol)}\par\smallskip
Public opinion on government-provided health insurance is sharply divided, with strong support stemming from beliefs that healthcare is a right and a government responsibility. However, significant opposition arises from concerns about cost, potential inefficiency, and worries that government involvement would negatively impact the quality of care. Some also \hm{question eligibility requirements} and express distrust of government competence and spending priorities.
\end{minipage}}
\caption{\textbf{A minority opinion covered only by the differences statement.} Gemma 3 12B received the same thirty DeliberationBank opinions in the same order at temperature zero, once with each prompt. Sol labeled the opinion at the top as covered only by the statement on the right, based on the shaded reference to eligibility requirements, which does not mention citizenship. The statements are verbatim; the example was chosen by a fixed rule described below.}
\Description{A box with a minority opinion about citizenship as a condition for healthcare, and below it two statements: the agreement statement does not mention it, the differences statement mentions eligibility requirements, shaded.}
\label{fig:rq3example}
\end{figure}

\emph{Worked example.} The example in \figref{fig:rq3example} was chosen by a rule fixed before reading any statement: an open-weight mediator, order draw 0, the two plain prompts, both statements of two to four sentences, an agreement-seeking statement covering at least one higher-support position, and exactly one minority position covered by the difference-setting statement but not by the agreement-seeking one; we took the first qualifying question in dataset order. In this task, five of the thirty opinions oppose government health insurance. Sol labels both statements as covering four of these opinions and only the differences statement as covering the fifth, through a broad reference to eligibility requirements that leaves the citizenship condition unspecified. In 90\% of matched pairs on DeliberationBank sample~1, Sol labels at least one minority opinion as covered under differences and omitted under agreement. Pairs match the question, mediator, opinion order and concreteness setting, across the four main mediators. Questions contain between two and nineteen minority positions, so this is a lenient criterion; the contrast in \figref{fig:objective}c is the quantitative measure.

The four comparable prompts provide the main objective contrast; adding legislative compromise and irreducible conflict prompts changes the comparison. Table~\ref{tab:representationgapintervals} retains both scopes. In the main comparison, DB sample 1 and Polis use Qwen3, Gemma 3, Mistral and Terra; DB samples 2 and 3 use Qwen3, Gemma 3, Llama 3.1 and Phi-4. Each judge uses its own DeliberationBank opinion summaries and stance labels; Polis comments are used verbatim and their groups come from votes. Thus the two judge pipelines compare measurement choices as well as coverage readings. Table~\ref{tab:forest} gives the equal-question objective contrast for every sample and judge, and Table~\ref{tab:factorialjudges} the corresponding logistic GEE interaction with standard and bias-reduced inference.

\emph{Alternative explanations.} Adjusting the Sol interaction for log statement length leaves the odds ratio at 0.249 (unadjusted 0.251) on DeliberationBank sample~1 and 0.080 (0.080) on Polis. Under agreement-seeking, the gap is 22.6 percentage points without and 29.5 with the concreteness request on DeliberationBank, and 18.2 and 24.6 on Polis. Among agreement-seeking statements that cover at least the median number of positions of difference-setting statements (21 on DeliberationBank, 10 on Polis), the gap is 26.2 percentage points ($n=68$ statements) and 43.5 ($n=39$), against 26.1 and 21.4 for all agreement-seeking statements. Across 4 group definitions, the equal-question contrast ranges from 37.9 to 42.6 percentage points on DeliberationBank sample~1 and from 19.6 to 26.9 on Polis. Sol and Luna put 92.3\%, 93.0\% and 84.3\% of DeliberationBank positions in the same group in samples 1--3. Because each judge produces its own stance labels, the questions that contain both groups depend on the judge (Table~\ref{tab:forest}, column $q$). Under Sol, no stance reaches 40\% in tipping (samples 1 and 2) and influencers (sample 3), and one question in sample 3 has no minority position; such questions drop out of the equal-question contrast, while pooled gaps keep them.

\emph{Several agreement statements together.} For each matched cell (question, mediator and concreteness setting), we count a minority position as covered by the five agreement statements of the order reruns if at least one of them covers it, and compare this with the mean coverage of a single differences statement from the same cell. Questions receive equal weight and only questions with both groups are used. Under Sol, the five agreement statements together cover 33.8\% of minority positions on DeliberationBank sample~1, 39.1\% on sample~3 and 8.6\% on Polis; a single differences statement covers 72.1\%, 75.7\% and 32.1\% (gaps of 38.3, 36.5 and 23.6 percentage points; 95\% CI 23.6--51.8, 22.0--50.5 and 16.7--30.4). A single agreement statement covers 16.7\%, 25.2\% and 3.2\%. Under Luna, the corresponding union and single-differences values are 58.3\% and 86.9\% (sample~1), 62.2\% and 90.7\% (sample~3) and 29.3\% and 47.9\% (Polis), with gaps of 28.7 (13.4--44.1), 28.5 (18.4--38.4) and 18.6 (11.6--27.2) points.

\emph{Coverage across reruns.} For coverage, Sol read 125 DeliberationBank and 31 Polis statements twice. For each position in these statements, we compare the change in its coverage label between the two readings with the change between the statement and the four other order reruns of the same task, averaging within question and weighting questions equally. The label changed in 8.5\% of cases on rereading and 18.8\% between orders on DeliberationBank (excess 10.3 percentage points, 95\% CI 8.4--12.3), and in 3.1\% and 17.1\% on Polis (14.0, 11.6--16.6). Luna's rereadings are less consistent, so the excess is smaller under Luna.

\emph{Polis self-descriptions.} Some Polis conversations include comments in which participants describe themselves (``I am under 30 years of age'', ``I live in a rural area'') and which others vote on. A regular expression for first-person statements of age, place of residence, gender, income, education, housing and occupation identifies 29 such comments among the sampled ones, 19 of them minority positions and 6 higher-support positions. These comments state no view on the policy, so a statement about the policy has little reason to cover them. Excluding them leaves the gap under the agreement prompts nearly unchanged (20.0 percentage points, against 20.1) and changes the gap under the differences prompts from \textminus 1.7 to \textminus 24.1 points (95\% CI \textminus 39.0 to \textminus 9.3): without them, the differences prompts cover minority positions more often. The objective contrast rises from 21.8 to 44.1 points under Sol (28.9--59.9) and from 24.4 to 49.3 under Luna (37.0--62.1). A narrower rule that excludes the 25 comments beginning with ``I'' gives 42.3 and 46.8 points. The main analysis keeps all comments.

\begin{table}[!htb]\centering\footnotesize
\caption{\textbf{The objective contrast by sample and judge.} Higher-support-minus-minority coverage gap under agreement-seeking minus the gap under difference-setting, in percentage points, four factorial prompts. Questions are weighted equally and restricted to those containing both groups ($q$). Intervals resample questions; LOO gives the range when one question is left out. Haiku reads Sol\textquoteright{}s position summaries and groups.}
\label{tab:forest}
\begin{tabular}{@{}llrrrrrl@{}}\toprule
Sample & Judge & $q$ & Statements & Agreement gap & Differences gap & Contrast (95\% CI) & LOO range \\ \midrule
DB sample 1 & Sol & 9 & 800 & 34.5 & $-$3.4 & 37.9 (19.7--55.5) & 32.4--43.4 \\
DB sample 2 & Sol & 9 & 160 & 22.0 & $-$6.0 & 28.0 (7.8--51.3) & 20.7--32.7 \\
DB sample 3 & Sol & 8 & 800 & 28.4 & $-$9.9 & 38.3 (14.6--57.5) & 33.2--48.3 \\
Polis & Sol & 9 & 800 & 20.1 & $-$1.7 & 21.8 (14.8--29.7) & 19.0--23.3 \\
DB sample 1 & Luna & 9 & 800 & 40.9 & 3.0 & 37.9 (22.2--53.7) & 32.9--42.8 \\
DB sample 2 & Luna & 9 & 160 & 23.0 & $-$6.0 & 29.0 (15.8--44.9) & 23.0--32.4 \\
DB sample 3 & Luna & 10 & 800 & 24.9 & $-$3.9 & 28.8 (13.7--42.7) & 24.7--34.7 \\
Polis & Luna & 9 & 800 & 33.9 & 9.5 & 24.4 (17.4--31.3) & 22.5--26.9 \\
DB sample 1 & Haiku & 9 & 800 & 16.8 & $-$21.2 & 37.9 (20.6--55.2) & 32.5--43.5 \\
Polis & Haiku & 9 & 800 & 18.9 & $-$0.5 & 19.3 (12.8--27.0) & 16.4--20.8 \\
\bottomrule\end{tabular}\end{table}

\begin{table}[!htb]\centering\footnotesize
\caption{\textbf{Higher-support minus minority coverage, in percentage points}, pooled over positions, for the four comparable prompts (as in \figref{fig:objective}a, b) and for all six prompts, which add the legislative compromise and irreducible conflict prompts. 95\% intervals resample questions and do not include uncertainty in the LLM-judge labels.}
\label{tab:representationgapintervals}
\begin{tabular}{@{}llrr@{}}\toprule
Sample & Objective & Four prompts & Six prompts \\ \midrule
DB sample 1 & Find agreement & +26.1 (+8.6--+44.3) & +20.5 (+4.0--+37.7) \\
DB sample 1 & Set out differences & $-$5.0 ($-$17.2--+8.5) & $-$2.0 ($-$13.6--+10.9) \\
Polis & Find agreement & +21.4 (+16.4--+25.9) & +24.9 (+18.3--+30.8) \\
Polis & Set out differences & $-$1.2 ($-$11.7--+8.2) & +0.0 ($-$9.8--+7.9) \\
DB sample 2 & Find agreement & +21.6 (+8.3--+34.4) & +19.7 (+9.0--+30.9) \\
DB sample 2 & Set out differences & $-$3.8 ($-$18.5--+8.2) & $-$2.4 ($-$16.0--+9.2) \\
DB sample 3 & Find agreement & +29.1 (+1.1--+50.3) & +25.6 ($-$0.1--+46.7) \\
DB sample 3 & Set out differences & $-$5.3 ($-$19.2--+6.9) & $-$4.7 ($-$20.0--+7.6) \\
\bottomrule\end{tabular}\end{table}

\begin{table}[!htb]\centering\footnotesize
\caption{\textbf{Interaction between minority status and the agreement objective.} Each judge uses its own opinion summaries and stance labels. An odds ratio below 1 indicates that the relative odds of coverage for minority versus higher-support positions are lower under agreement than under differences. Tests are one-sided, with standard robust and bias-reduced standard errors. ``Same direction'' counts mediator models with a larger coverage gap under agreement.}
\label{tab:factorialjudges}
\begin{tabular}{@{}llllrr@{}}\toprule
Sample & LLM judge & OR (95\% CI) & $p$ & $p$, bias-reduced & Same direction \\ \midrule
DB sample 2 & Sol & 0.34 (0.15--0.75) & 0.0038 & 0.0101 & 4 of 4 \\
DB sample 3 & Sol & 0.21 (0.08--0.56) & 0.0009 & 0.0030 & 4 of 4 \\
DB sample 2 & Luna & 0.27 (0.11--0.66) & 0.0021 & 0.0082 & 4 of 4 \\
DB sample 3 & Luna & 0.22 (0.12--0.40) & $4.5\times 10^{-7}$ & $6.9\times 10^{-6}$ & 4 of 4 \\
\bottomrule\end{tabular}\end{table}

Table~\ref{tab:coverage} compares four conditions: the baseline prompt, an intermediate list of opinions, an instruction to ``reflect the substance of every citizen's view'', and both additions together. In the list conditions, Phi-4 prepares one-line summaries of the opinions before one of the three main open-weight mediators writes the proposal. Coverage increases alongside output length; this does not identify a causal mechanism. The minority-by-instruction interaction odds ratio is 0.82 (95\% CI 0.67--1.00) on DeliberationBank and 0.62 (0.52--0.74) on Polis. Values below 1 indicate a smaller multiplicative change in coverage odds for minority positions than for higher-support positions. Recovery of explicitly planted sentences is 86\% for minority and 88\% for higher-support sentences. This supports detection of explicit content, but does not validate paraphrased coverage or establish endorsement and influence.

\begin{table}[!htb]
\centering
\footnotesize
\caption{\textbf{The instruction to cover everyone} (three models, two conditions that request concreteness, five orders at $T{=}0$; minority positions by stance on DeliberationBank and by participants' votes on Polis). Represented positions (\%), gap in percentage points, mean words, share of statements that assign someone and decoy false-positive rate.}
\label{tab:coverage}
\resizebox{\ifdim\width>\textwidth\textwidth\else\width\fi}{!}{%
\begin{tabular}{@{}llrrrrrr@{}}
\toprule
Data & Arm & Higher support & Minority & Gap & Words & Assigning (\%) & Decoys (\%) \\
\midrule
DeliberationBank & Baseline & 61 & 48 & 12.6 & 88 & 29 & 0.07 \\
 & List of views & 59 & 48 & 11.3 & 91 & 31 & 0.07 \\
 & List + instruction & 67 & 54 & 12.6 & 114 & 33 & 0.07 \\
 & Instruction & 69 & 53 & 16.6 & 112 & 34 & 0.07 \\
Polis & Baseline & 29 & 19 & 10.4 & 99 & 37 & 0.40 \\
 & List of views & 29 & 18 & 11.0 & 102 & 35 & 0.20 \\
 & List + instruction & 47 & 24 & 22.4 & 135 & 40 & 0.13 \\
 & Instruction & 45 & 22 & 22.4 & 128 & 42 & 0.13 \\
\bottomrule
\end{tabular}}
\end{table}

\end{document}